\documentclass[final,3p,twocolumn,authoryear,sort&compress,times]{maia/maia}
\journal{MAIA Master}

\usepackage{amsmath,amssymb}	
\usepackage{graphicx}
\usepackage{natbib}
\usepackage{bbm}
\usepackage{multirow}
\usepackage{booktabs}
\usepackage[T1]{fontenc}

\begin{document}
\begin{frontmatter}

\title{Live(r) Die: Predicting Survival in Colorectal Liver Metastasis}

\author[MBP,SRI]{Muhammad Alberb}
\author[MI,SRI]{Helen Cheung}
\author[MBP,SRI]{Anne Martel}
\address[MBP]{Department of Medical Biophysics, University of Toronto, Toronto, ON, Canada}
\address[MI]{Department of Medical Imaging, University of Toronto, Toronto, ON, Canada}
\address[SRI]{Sunnybrook Research Institute, Toronto, ON, Canada}

\begin{abstract}
Colorectal cancer frequently metastasizes to the liver, significantly reducing long-term survival. While surgical resection is the only potentially curative treatment for colorectal liver metastasis (CRLM), patient outcomes vary widely depending on tumor characteristics along with clinical and genomic factors. Current prognostic models, often based on limited clinical or molecular features, lack sufficient predictive power, especially in multifocal CRLM cases. In this work, we present a fully automated framework for surgical outcome prediction from pre- and post-contrast MRI acquired before surgery. Our framework consists of a segmentation pipeline and a radiomics pipeline. The segmentation pipeline learns to segment the liver, tumors, and spleen from partially annotated data by leveraging promptable foundation models to complete missing labels. Moreover, we propose SAMONAI, a novel zero-shot 3D prompt propagation algorithm that leverages the Segment Anything Model (SAM) to segment 3D regions of interest from a single point prompt, significantly improving our segmentation pipeline's accuracy and efficiency. The predicted pre- and post-contrast segmentations are then fed into our radiomics pipeline, which extracts features from each tumor and predicts survival using SurvAMINN, a novel autoencoder-based multiple instance neural network for survival analysis. SurvAMINN jointly learns dimensionality reduction and hazard prediction from right-censored survival data, focusing on the most aggressive tumors. Extensive evaluation on an institutional dataset comprising 227 patients demonstrates that our framework surpasses existing clinical and genomic biomarkers, delivering a C-index improvement exceeding 10\%. Our results demonstrate the potential of integrating automated segmentation algorithms and radiomics-based survival analysis to deliver accurate, annotation-efficient, and interpretable outcome prediction in CRLM.
\end{abstract}


\end{frontmatter}

\section{Introduction}
\label{sec:introduction}
Colorectal cancer (CRC) is the development of cancerous cells that originate in the colon or the rectum. It is the third most common form of cancer and the second leading cause of cancer-related death worldwide \citep{Sung2021GLOBOCAN}. Primary CRC often metastasizes to other organs, forming secondary tumors. The liver is the most common site of CRC metastasis, due to its anatomical and physiological characteristics \citep{Wang2023CRLM}. In particular, the blood supply of the liver is directly connected to the colon through the portal vein, which facilitates metastasis of cancerous cells to the liver without systemic spread elsewhere \citep{Zhou2022LiverMetastasis}. While CRC is associated with a relatively high survival rate when non-metastatic, the development of metastases significantly reduces long-term survival \citep{Rawla2019CRC}.

Surgical resection is the gold standard treatment for colorectal liver metastasis (CRLM)  and is considered the only potentially curative strategy \citep{Dong2023Surg}. However, surgery is not suitable for every CRLM patient. Postoperative outcomes highly depend on tumor characteristics along with clinical and molecular factors \citep{Spolverato2013SurgicalOutcome, Kuo2015TumorLocation}. Accurately predicting surgical outcome is essential to avoid non-beneficial surgeries, allow for personalized therapy, and ultimately enhance survival outcomes for patients with CRLM \citep{Margonis2022PrecisionSurgery}. 

Conventional prognostic scores are based on different clinical and molecular factors \citep{Tian2024PrognosticFactors}. However, none of these are strong predictors of long-term survival \citep{Roberts2014BiomarkersLimitations, Margonis2022PrecisionSurgery}. Consequently, various imaging-based scores have been proposed to predict CRLM patient outcomes \citep{Cheung2019TTE, Nakai2020CRLMPeripheral, Sasaki2018TBS}. In cases involving multiple metastases, prognosis is often driven by the most aggressive tumors, which serve as stronger indicators of patient outcomes. \citep{Bahbahani2023TumorAggression}. However, since it is difficult to identify how each tumor contributes to prognosis, those imaging risk scores focus mainly on the largest tumors. As a result, they tend to overlook the potential impact of multifocal CRLM, where multiple tumors are present \citep{Chen2021AMINN}.

Another limitation of those imaging scores is their reliance on manual tumor segmentation performed by experts, which is prone to inter-rater variability \citep{Cheung2019TTE, Sasaki2018TBS}. In addition, the annotation process is both time-consuming and labor-intensive, particularly for three-dimensional imaging modalities such as computed tomography (CT) and magnetic resonance imaging (MRI) \citep{Horvat2024RadiomicsLimitations}. While deep learning has been widely adopted to automate medical image segmentation, training models using supervised methods remains constrained by the scarcity of fully-annotated datasets. Moreover, when segmentation outputs are used to extract features for downstream models, such as those for outcome prediction, inaccuracies can significantly compromise prognostic performance \citep{Chen2021AMINN}.

A possible explanation is that existing automatic segmentation and feature extraction methods used in prognostic studies often rely solely on post-contrast images, which are acquired after the injection of contrast agents to improve tissue and lesion visibility \citep{Chen2019AEGMM, Chen2021AMINN}. However, in clinical practice, radiologists typically evaluate both pre- and post-contrast scans to accurately detect lesions and assess their malignancy \citep{Chernyak2018LIRADS}. This discrepancy underscores the importance of incorporating pre-contrast imaging information into segmentation-based prognostic models.

As outlined above, current CRLM prognostic methods exhibit several key limitations: (1) a tendency to overlook multifocality, (2) reliance solely on post-contrast images, and (3) dependence on manual segmentation masks or, alternatively, (4) the need for fully annotated datasets to enable automated segmentation.

\begin{figure*}[tb]
    \centering
    \includegraphics[width=\textwidth]{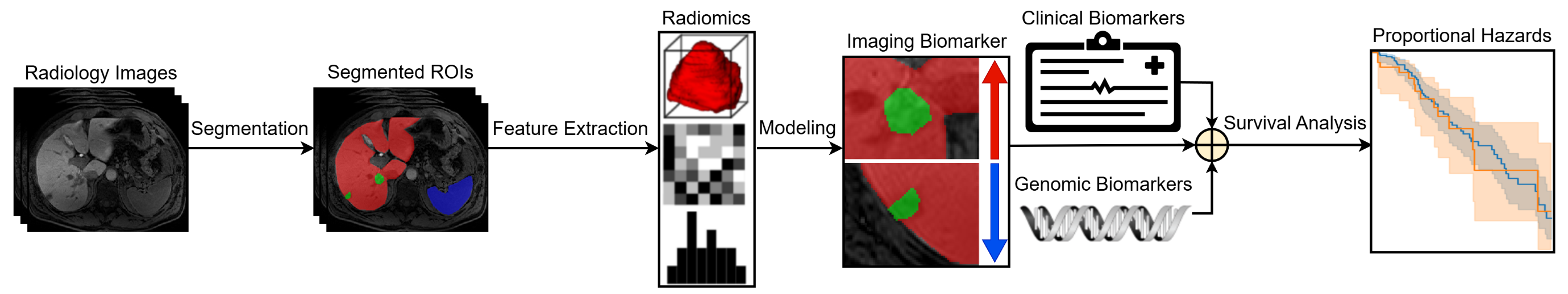}
    \caption{Framework overview. Liver, tumors, and spleen are segmented and used to extract radiomic features. An imaging risk score is then predicted from radiomics and combined with other biomarkers to predict patient prognosis.}
    \label{fig:overview}
\end{figure*}

To this end, we propose an end-to-end fully automated, multifocal, and annotation-efficient framework for predicting CRLM surgical outcomes from pre- and post-contrast MRI. Specifically, our framework is composed of a segmentation pipeline followed by a radiomics pipeline. The segmentation pipeline learns to segment the liver, liver tumors, and spleen from partially-annotated non-contrast as well as contrast-enhanced MRI. Afterwards, the radiomics pipeline extracts multifocal radiomic features from the acquired segmentations and learns to predict survival based on the most aggressive tumors. An overview of our proposed framework is illustrated in figure~\ref{fig:overview}.

The rest of this document is organized as follows: In section~\ref{sec:state_of_the_art}, we conduct a review of the state of the art, highlighting latest advancements and current limitations. In section~\ref{sec:material_and_methods}, we describe our dataset as well as the proposed framework in detail. Section~\ref{sec:results} shows the results of our approach and how it compares with existing methods. In Section~\ref{sec:discussion}, we present a thorough analysis of the results and discuss the strengths of our framework as well as potential areas for improvement. Finally, the conclusions of this study are presented in section~\ref{sec:conclusions}.
\section{State of the art}
\label{sec:state_of_the_art}
Medical image segmentation and radiomics-based surgical outcome prediction have been hot topics in recent years. This popularity is attributed to their critical roles in clinical decision-making and patient management. To this end, we review the existing literature, summarizing key methodologies, prevailing trends, and common limitations, with a particular focus on their application to CRLM.

\subsection{Segmentation}

Since its debut, deep learning has cemented itself as the mainstream solution for automatic segmentation. Pioneering this direction, \citet{Ronneberger2015UNet} introduced UNet, an encoder-decoder convolutional neural network (CNN) supported with skip connections that enables precise segmentation. UNet, which quickly became a benchmark model, paved the way for two major research directions: architectural innovation and framework design. For instance, \citet{Isensee2021nnUNet} proposed nn-UNet, a highly versatile UNet-based framework that automatically adapts its pipeline to different datasets and tasks. On the other hand, after \citet{Dosovitskiy2021ViT} introduced the vision transformer (ViT), several transformer-based segmentation architectures, such as UNETR \citep{Hatamizadeh2022UNETR} and Swin-UNETR \citep{Hatamizadeh2022SwinUNETR}, were proposed. Thanks to the higher learning capacity of the attention mechanism compared to convolution \citep{Vaswani2017attention}, those transformer-based architectures have the potential to out-scale CNN-based ones in abundant data scenarios. 

Despite demonstrating their effectiveness in different segmentation applications, these methods heavily rely on the availability of a large, homogeneous, and precisely annotated dataset \citep{Rayed2024DLLimitations}. More often than not, these requirements are difficult to achieve in real clinical settings. Therefore, alternative approaches with looser requirements have emerged. For instance, the concept of transfer learning has widely gained popularity. Transfer learning aims to leverage the knowledge gained by learning one task to facilitate the learning of another \citep{Zhao2024TransferLearning}. For example, pretraining architectures on natural image classification is a common practice. Afterwards, the learned weights are used instead of random initialization when applied to a new task, leading to a faster and a more stable convergence. However, this classical pretraining method relies on the availability of labels and therefore fails to leverage unannotated data. 

Recently, several self-supervised pretraining strategies have been proposed to overcome this limitation. Self-supervised learning is a pretraining paradigm where models are trained using the data itself, rather than relying on externally provided labels \citep{Zhao2024TransferLearning}. One common example is masked language or vision modeling, which involves training models to predict missing parts of a masked input \citep{He2022MAE}. Contrastive learning is another approach that trains models to distinguish between similar and dissimilar inputs \citep{Chen2020SimCLR}. Most contrastive learning approaches treat augmented views of the same input as similar, and entirely different inputs as dissimilar. This assumption allows for self-supervised contrastive pretraining that does not require external labels. Those pretraining strategies are the main engine fueling the current foundation modeling trend. Foundation models are created by training highly-parameterized models on vast amounts of data to be applied to a wide range of tasks \citep{Moor2023FoundationModels}. When applied to medical images, foundation models can sometimes have acceptable zero-shot performance, which is performance without providing any annotated samples. Additionally, few-shot learning can significantly improve their performance by finetuning over a few annotated samples \citep{Woerner2025ZeroVsFewShot}. 

Although foundation models tend to generalize over a wide range of tasks, they are limited to predetermined input types. For example, 2D foundation models are not inherently capable of processing 3D volumes. Since many medical imaging modalities, such as MRI, produce 3D volumes, image pretrained foundation models are typically applied independently to individual slices, resulting in the loss of crucial depth information. Recently, a few studies addressed this issue by proposing 3D foundation models. For instance, \citet{Tang2022FoundationSwin} pretrained a Swin-UNETR on 5050 CT volumes. Swin Transformers, proposed by \citet{Liu2021SwinTransformer}, extract hierarchical features suitable for dense vision tasks, allowing them to surpass vanilla ViTs in segmentation. Although Swin Transformers are the current state-of-the-art in segmentation, the formulation of ViT enables several pretraining strategies and optimizations. Therefore, \citet{xu20253DINO} adopted the ViT architecture in their 3D SSL method, yielding 3DINO, a 3D multimodal foundation model trained on $\sim$100,000 volumes.

While foundation models successfully address the problem of relying on large datasets to achieve generalizability, finetuning them for specific tasks still requires a fully annotated dataset, which is usually difficult to obtain in clinical environments. Therefore, a few research directions addressing this problem emerged. For instance, \citet{Ulrich2023MultiTalent} proposed a method to learn from multiple partially-annotated CT datasets based on a class-adaptive loss function. However, this approach assumes the availability of multiple homogeneous datasets with annotations for different targets. This might be a valid assumption for CT. However, MRI tends to produce volumes that are significantly different depending on the used configurations and contrast-agents. Therefore, if MRI datasets from different sources with different segmentation targets are combined, a model can learn to distinguish between data sources rather than learning to jointly segment the intended targets. A possible alternative is to create fully-annotated datasets by segmenting the missing targets using a general purpose method, such as promptable foundation models.

Promptable models, which are controllable via user input prompts, tend to be more generalizable than their non-promptable counterparts \citep{Zhou2022PromptGeneralization}. One of the leading promptable foundation models for segmentation, proposed by \citet{Kirillov2023SAM}, is the Segment Anything Model (SAM). SAM uses an image and a prompt encoder to extract features from the images as well as the regions of interest based on different types of user inputs, such as points, bounding boxes, and masks. A lightweight mask decoder then combines image and prompt encodings to generate final segmentation masks. By training on a massive dataset of over 1 billion masks across 11 million images, SAM learns a general definition of an object and is able to generalize over unseen targets and domains. Since zero-shot performance can usually be improved by domain-specific training, \citet{Ma2024MedSAM} proposed MedSAM, a medical image pretrained version of SAM that uses bounding box prompting. According to their evaluation, MedSAM is able to outperform SAM in most tasks. Therefore, it has been widely adopted for prompt-based medical image segmentation. Despite the release of SAM2 \citep{Ravi2024SAM2} and MedSAM2 \citep{Ma2025MedSAM2} which are able to segment 3D medical images by treating them as video frames, several studies reported that they fail to consistently outperform SAM and MedSAM, especially for liver segmentation \citep{Zhu2024MedicalSAM2, Sengupta2025SAMvsSAM2}.

MedSAM offers numerous advantages. However, it still has a few limitations. For example, it is not capable of processing different types of prompts. While box prompting can be less ambiguous and suitable for most medical image segmentation tasks, it is not a good fit for large and concave-shaped organs, such as the liver and the spleen. For example, fitting a bounding box over the liver overlaps with most abdominal organs. The 2D nature of MedSAM is another limitation, since it ideally requires a prompt per slice. \citet{Isensee2025nnInteractive} have recently addressed those limitations by proposing nnInteractive, a 3D promptable foundation model for medical image segmentation. By adding prompts as extra channels, nnInteractive simply trains nnUNet as a promptable model. The 3D nature of nnInteractive does not only make it depth-aware but also allows for a single prompt per volume. Therefore, nnInteractive has the potential to outperform SAM-based methods in 3D medical image segmentation.

Although training promptable models on medical images can improve their performance, it comes at the cost of limited generalizability. The performance of these models is expected to drop when tested on unseen modalities. Therefore, utilizing these models might not be reliable for modalities such as abdominal MRI, since imaging characteristics can vary significantly depending on factors such as the type of sequence or contrast agent. Therefore, an extension of the original SAM that does not require explicit medical image finetuning can still be favorable if it addresses SAM's and MedSAM's major limitations of requiring a prompt per slice, lacking depth awareness, and relying on box prompting.

\subsection{Radiomics-based Outcome Prediction}

Radiomics, a term coined by \citet{Lambin2012Radiomics}, refers to the extraction of quantitative features from medical images. Radiomics leverages routinely acquired imaging data, such as CT or MRI, to identify patterns that may not be recognizable by human observers. Radiomic features typically include statistical, shape, and textural features. Recent advancements have introduced deep radiomics, a paradigm shift that incorporates features learned from deep neural networks rather than relying solely on hand-crafted descriptors \citep{Scapicchio2021DeepRadiomics}. Nevertheless, using predetermined descriptors improves interpretability and reduces the risk of overfitting, a key consideration when designing radiomics pipelines. Therefore, we focus on traditional radiomic features in this work, following several other recent studies \citep{Mariotti2025RadiomicsReview}.

Radiomic features have been widely used in oncology research for tasks such as tumor classification, treatment response prediction, and survival analysis \citep{Shur2021RadiomicsOncology}. The conventional radiomics pipeline involves image acquisition, region of interest segmentation, feature extraction, feature selection or dimensionality reduction, and modeling. However, the partially independent feature selection and decision-making stages often lead to loss of predictive features \citep{Liyanage2020JointSelectClassif}. 

To bridge this gap, \citet{Chen2021AMINN} proposed AMINN, an autoencoder-based multiple instance network that jointly learns dimensionality reduction of radiomics via an autoencoder branch and prediction of 3-years survival via a multilayer perceptron (MLP) branch. Additionally, AMINN considers multifocal CRLM by formulating the classification as a multiple instance learning (MIL) problem. MIL is a weakly supervised approach that links patient-level survival to tumor-level features by treating each tumor as an instance within a labeled patient-level bag. 

That said, there are a few limitations that can prohibit AMINN's real clinical application. For instance, treating the problem of survival prediction as a classification problem leads to loss of data points that were censored before the determined 3 years, which can be a significant subset that provides beneficial information. 

Moreover, despite considering multiple tumors, AMINN determines patient-level prediction based on the average of all of the tumors. In case of the existence of a high number of tumors, this aggregation could lead to a weaker signal from lethal tumors. Moreover, this could potentially be the reason AMINN's performance becomes unsatisfactory when using predicted segmentations for feature extraction rather than manual segmentations. These predicted segmentations generated by an automated segmentation model may contain numerous false positives, such as healthy liver parenchyma or vessels. These inaccuracies can dilute the signal of lethal tumors, even though such irrelevant regions are not supposed to influence the final outcome.

Another serious issue when using predicted segmentations for CRLM survival prediction is the possibility of having false positives that are not parts of the liver. Since decision-making models are not trained on features from non-liver regions, they are vulnerable to hallucination, potentially predicting bad scores for those regions. Therefore, it is extremely useful to develop anatomy-aware tumor segmentation algorithms that can exclude any non-liver predictions.

\subsection{Our Contributions}

In this work, we address the observed limitations of existing methods for surgical outcome prediction of CRLM. Specifically, our contributions can be summarized as:

\begin{itemize}
  \item We design a multi-class segmentation pipeline that can learn from partially-annotated data by leveraging promptable foundation models to complete missing annotations of the liver and spleen.
  \item We propose SAMONAI, a 3D prompt engineering strategy that leverages SAM's zero-shot performance to segment the liver and the spleen from a single point per object, introducing depth awareness and eliminating the need for a prompt per slice.
  \item We develop a multi-phasic radiomics pipeline for survival prediction of CRLM based on predicted segmentations.
  \item We design an autencoder-based multiple instance neural network for survival prediction (SurvAMINN) that jointly learns dimensionality reduction and prediction, makes decisions based on lethal tumors, and is capable of learning from censored data.
\end{itemize}
\section{Material and methods}
\label{sec:material_and_methods}

\subsection{Dataset}
\label{subsec:dataset}

The dataset used for this study was internally acquired at our institution (Sunnybrook Health Sciences Centre) between January 2013 and December 2020. The retrospective cohort consists of 227 CRLM patients who have received pre- and post-contrast T1-weighted MRI with the hepatobiliary contrast agent gadoxetic acid (Primovist, Eovist). Ethical approval for this retrospective study was obtained from the institutional research ethics board, which waived the requirement for individual informed consent.

Studies were obtained as per institutional clinical protocols. The patients received 3D axial T1 imaging (TE \textasciitilde 1.5 ms, TR \textasciitilde 3.0 ms, flip angle \textasciitilde 10 degrees) in the pre-contrast and hepatobiliary phases (20 minute delay).  For the contrast agent, a standard 10 mL intravenous dose of gadoxetate at 1.0 mmol/mL was administered.  Studies were performed on 1.5-T (GE Twinspeed\textsuperscript{\tiny{TM}}) or 3.0-T (Philips Achieva\textsuperscript{\tiny{TM}}) magnets with an eight-channel body phased array coil covering the liver.

The dataset also includes segmentations of CRLM tumors with a longest diameter $\geq 10\,\text{mm}$ for pre- and post-contrast images, overall mortality, and time to survival/death. The manual segmentations were conducted by an abdominal radiologist with 13 years of experience in abdominal segmentation. To validate unsupervised liver and spleen segmentation pipelines, the liver and the spleen were manually segmented by the same radiologist for 20 post-contrast images.

The number of segmented tumors per patient varies from 1 to 12 tumors, resulting in a total of 531 and an average of 2.4 tumors per patient. 130 patients have multifocal CRLM with an average of 3.4 tumors per patient. 82 patients have known postoperative survival times, with a median survival of 28 months. The rest of the patients are right-censored.

\begin{figure}[htb]
    \centering
    \includegraphics[width=\columnwidth]{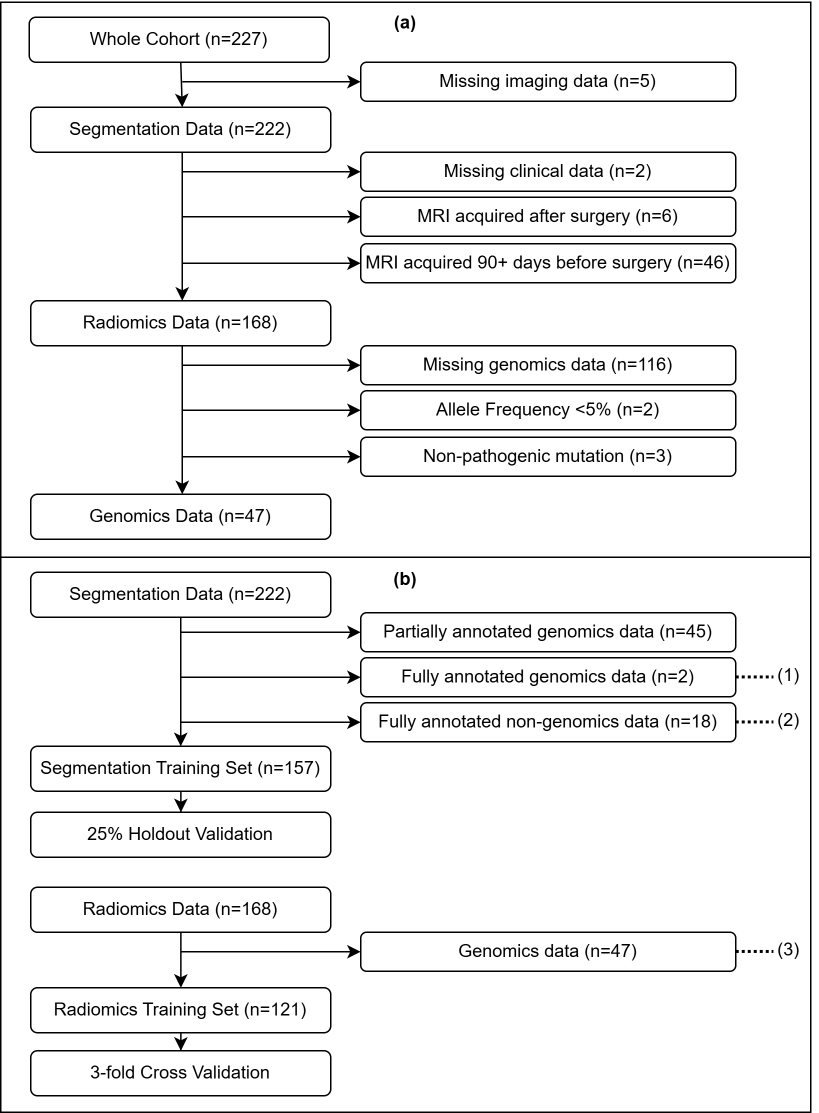}
    \caption{Data selection process. Different inclusion criteria for segmentation and outcome prediction result in different datasets. (a) Exclusion of cases that do not satisfy the selection criteria. (b) Datasets splitting for training, validation, and testing. For segmentation, a single holdout validation set was used due to computational constraints. Genomics data were excluded from both pipelines to be used as an independent testing set.}
    \label{fig:dataset}
\end{figure}

Furthermore, the dataset contains a few known prognostic variables. Specifically, the dataset includes demographic data such as age and sex, clinical biomarkers such as Fong score \citep{Fong1999Fong} and target tumor enhancement \citep{Cheung2019TTE}, as well as genomic biomarkers such as APC, TP53, KRAS, and NRAS mutational status available for a subset of patients. DNA extraction and mutation analysis were performed following the same procedure described by \citet{Seth2021MutationCRLM}. Genomic mutations were filtered based on their allele frequency and ClinVar pathogenicity classification, as described in figure~\ref{fig:dataset}. The number of occurrences of the specified mutations per patient was used as a prognostic biomarker.

For this study, different inclusion criteria were defined for the segmentation and outcome prediction pipelines. For segmentation, all patients with available pre- or post-contrast MRI with tumor segmentation were included. For outcome prediction, additional inclusion criteria were applied, such as available clinical data including survival information and preoperative MRI acquisition $\leq 3\,\text{months}$ prior to the surgery date. The subset with available liver and spleen segmentations was excluded from the segmentation pipeline, and the subset with genomic data was excluded for both pipelines to be used as independent testing sets. Figure~\ref{fig:dataset} summarizes the data inclusion criteria as well as splitting into training, validation, and testing sets.

\subsection{Segmentation Pipeline}

The goal of our segmentation pipeline is to segment the liver and CRLM tumors within it. The purpose of segmenting the liver is to exclude any predicted extra-hepatic tumors. Due to the similarity between the parenchyma of the liver and the spleen, liver segmentation algorithms often mistakenly segment parts of the spleen \citep{Hossain2023LiverSpleenAmbiguity}. Therefore, to reduce ambiguity and further improve the precision of tumor segmentation, we design our pipeline to segment the spleen as well.

\subsubsection{Segmentation with Partial Labels}
The primary challenge in developing an automatic segmentation algorithm for liver, tumors, and spleen lies in the absence of ground truth manual annotations for the liver and spleen. While tumor segmentations are fully available, manual annotations for liver and spleen were not initially performed due to their significantly larger size and consequently higher annotation cost compared to lesions. To address this, we propose a three-stage pipeline designed to learn effectively from partially annotated datasets.

In stage 1, we rely on a general-purpose semi-automated segmentation algorithm to fully annotate a random sample from the partially-annotated dataset. Specifically, we leverage promptable foundation models to generate labels for the liver and spleen. These labels are then combined with the available manual tumor labels. To minimize manual intervention, the sample size was limited to 20 cases.

In stage 2, we use the fully-annotated sample to fine-tune a fully-automated model. For this stage, the sample is split into 15 training and 5 validation cases. Due to the small sample size, training a deep neural network from scratch is associated with a high risk of overfitting, especially since the labels acquired in stage 1 are noisy. Therefore, we finetune a few-shot UNETR model using the pretrained ViT from 3DINO~\citep{xu20253DINO}, effectively reducing the number of trainable parameters. Due to the weak prior assumptions, 3DINO's plain ViT suffers from inferior performance on dense prediction tasks. Therefore, finetuning 3DINO leverages a ViT Adapter, originally proposed by \citet{Chen2022ViTAdapter}, that injects spatial inductive biases into the pretrained ViT using a simple CNN, improving performance on downstream segmentation tasks.

\begin{figure*}[tb]
    \centering
    \includegraphics[width=\textwidth]{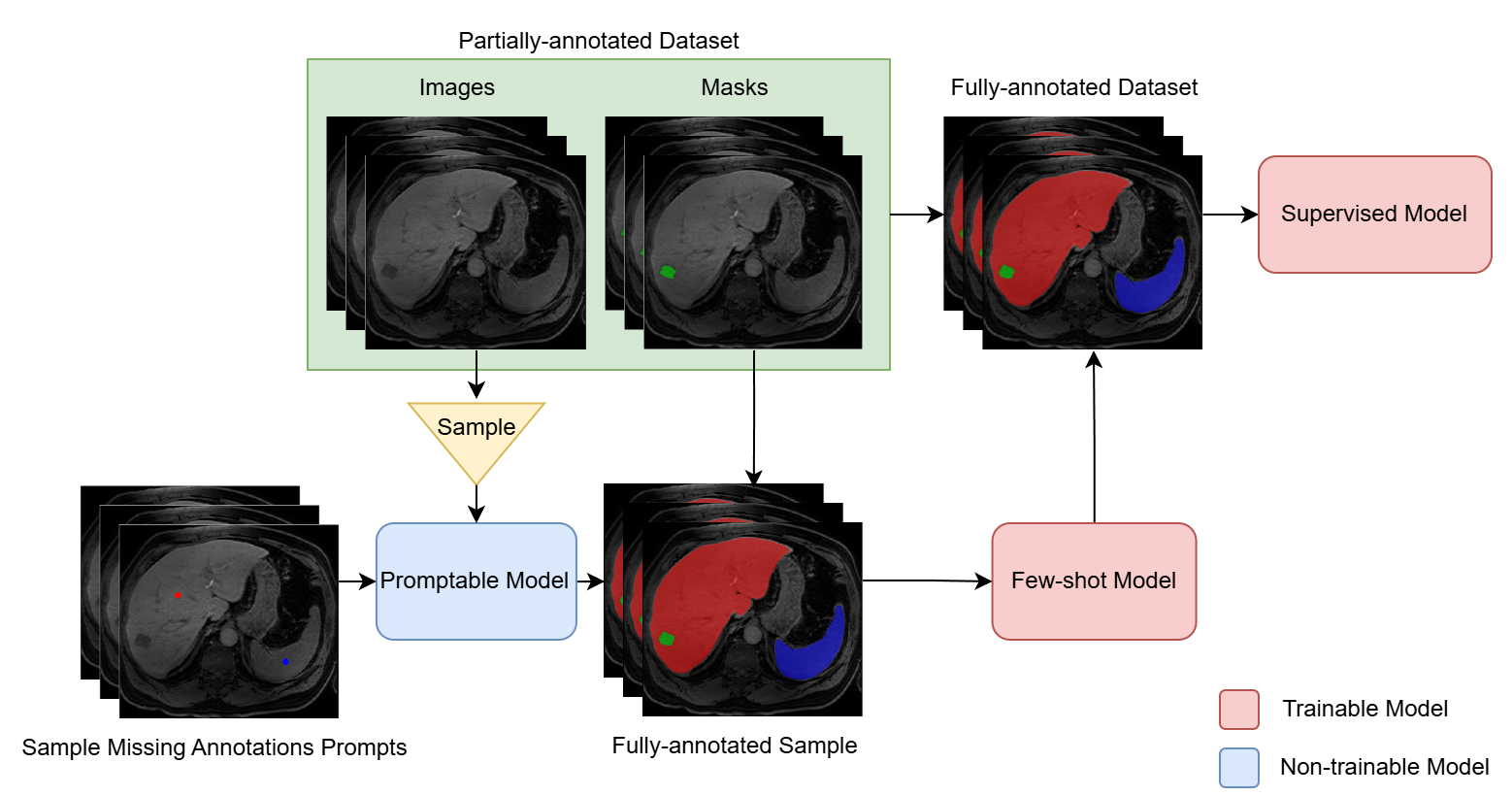}
    \caption{Segmentation pipeline. A promptable foundation model is used to complete partially-annotated masks for a small subset. A few-shot model is trained on the subset then used to complete the annotation for the whole dataset. Finally, A supervised model is trained on the whole fully-annotated dataset.}
    \label{fig:segmentation_pipeline}
\end{figure*}

Stage 2 creates a fully-automated model. However, since this model was trained on a small subset of the dataset, it does not learn from all the available ground-truth annotations. Therefore, in stage 3, we use the few-shot model acquired from stage 2 to complete the missing liver and spleen annotations for the whole dataset. The fully-annotated dataset is then used to retrain 3DINO, leveraging more labels and consequently improving tumor segmentation. We apply our segmentation pipeline to pre- and post-contrast MRI. An overview of the pipeline is shown in figure~\ref{fig:segmentation_pipeline}.

\subsubsection{Segmentation with Prompts}

Our segmentation pipeline heavily relies on the accuracy and more importantly the generalizability of the promptable model used in stage 1. Consequently, this model is the primary source of knowledge for targets with missing labels. The performance of existing medical imaging models, such as MedSAM, significantly varies based on the modalities included in their training data. Therefore, to ensure stable performance regardless of the modality or the contrast agent, we aim to use a general-purpose method, such as SAM, that is not particularly trained on medical images.

Since SAM is a 2D model originally proposed for natural images, we design SAMONAI, a simple yet powerful 3D zero-shot prompt propagation algorithm capable of segmenting entire 3D objects from a single point prompt per object. SAMONAI leverages SAM's predicted segmentation from one slice to generate candidate prompts then selects the ideal prompt according to predetermined prompt engineering criteria.

\begin{figure}[htb]
    \centering
    \includegraphics[width=\columnwidth]{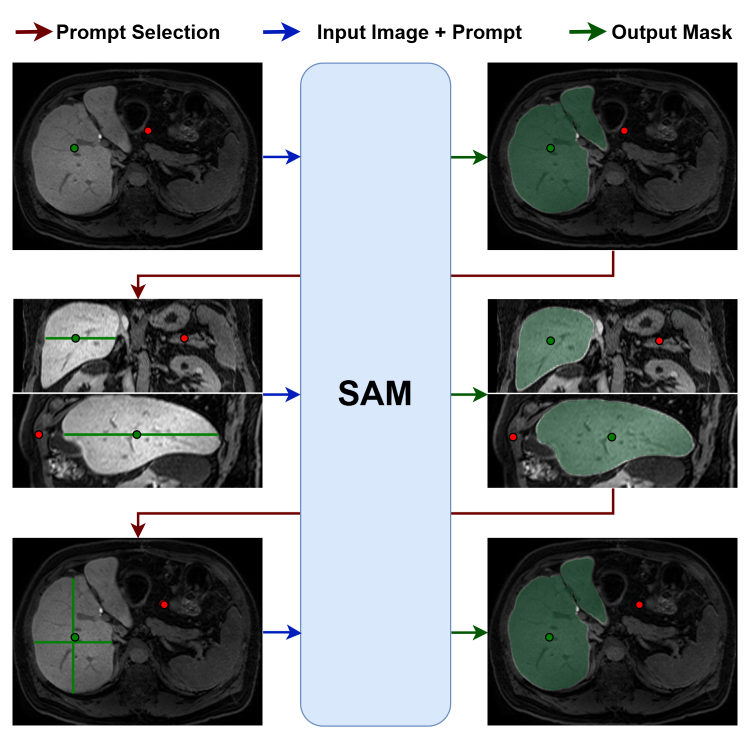}
    \caption{Overview of SAMONAI when applied to the liver. 3D objects are segmented from a single 2D point by propagating prompts to other slices over three steps. Positive and negative points are displayed in green and red, respectively.}
    \label{fig:samonai}
\end{figure}

SAMONAI operates over three steps, as shown in figure~\ref{fig:samonai}. Initially, the user defines a positive point in a single slice from any view for each object of interest. Additionally, a negative point for the background is optional to reduce ambiguity. SAMONAI uses SAM to segment the objects of interest in the slice and the view (e.g., axial) in which the points were added. 

This segmented slice from the first view appears in the two other views as a line of positive points in every slice that intersects with the segmented region. In the second step, SAMONAI segments a single slice from each of the two other views (e.g., sagittal and coronal). This slice is selected as the slice with the maximum number of positive points (longest line of positive points). A positive prompt for this slice is generated by selecting a point from the pool of positive points, and a negative prompt is selected from the pool of negative points along the line of positive points.

Segmenting an axial, a sagittal, and a coronal slice, in the first and second steps, creates three perpendicular lines of positive points covering the object of interest. In the third step, SAMONAI iterates over each slice, selects a positive prompt from the available positive points, and a negative prompt from the available negative points located outside a 3D bounding box fitted over the 3 perpendicular positive lines. The selected prompts are subsequently utilized to perform slice-by-slice segmentation. As all prompts are precomputed in advance, step 3 enables parallel execution of the segmentation process across available computational resources, thereby enhancing overall efficiency.

To introduce 3D awareness, this step is applied over each of the 3 views and the reconstructed 3D probability maps are averaged. To limit additional computational costs, slices from each view are sampled at a rate of $\frac{1}{3}$, and the missing slices are linearly interpolated. This approach preserves the computational cost of segmenting a single view while still incorporating 3D information. Finally, the logits of each object $L$ are binarized using an adaptive threshold $T_L$ defined as:
\[T_L = \mu_L + 2\sigma_L,\]
where $\mu_L$ and $\sigma_L$ are the mean and standard deviation of the logits $L$, respectively. This relatively high threshold is intended to reduce false positives, in line with our goal of excluding extra-hepatic tumors.

SAMONAI's second and third steps involve selecting an optimal point from a set of candidate points. Since random selection can return misleading prompts, we define a set of intuitive prompt engineering criteria to select the most promising point prompt. Our predefined criteria are based on the observation that SAM tends to generate better segmentations when:
\begin{enumerate}
    \item Prompts are close to the center of the object (i.e., away from the outer edges or boundaries).
    \item Prompts have an intensity similar to that of the object of interest (i.e., not too bright or too dark).
    \item Prompts are located in a homogeneous region (i.e., away from areas with high noise or sharp intensity transitions).
\end{enumerate}

\noindent Based on this, we define a location criterion $C_{l}$, an intensity criterion $C_{i}$, and a homogeneity criterion $C_{h}$. For a given point with coordinates $p$ from a set of candidate points with a set of coordinates $P$, the location criterion $C_{l}$ is defined as the Euclidean distance from the centroid of all candidate points, formulated as:

\begin{equation}
C_{l}(p) = \left\| p - \frac{1}{|P|} \sum_{q \in P} q \right\|
\label{eq:location}
\end{equation}

\noindent The intensity criterion $C_{i}$ is defined as the absolute difference between the intensity of the input image $I$ at point $p$ and the median intensity of all candidate points $P$, which can be formulated as:

\begin{equation}
C_{i}(p) = \left| I(p) - \text{median}_{q \in P}(I(q)) \right|
\label{eq:intensity}
\end{equation}

\noindent The homogeneity criterion $C_{h}$ is defined as the standard deviation of intensities within an $11 \times 11$ neighborhood centered at point $p$. Let $N_{11 \times 11}(p)$ denote the set of coordinates in the neighborhood. Then $C_h$ can be defined as:

\begin{equation}
C_{h}(p) = \sqrt{ \frac{1}{|N|} \sum_{r \in N_{11 \times 11}(p)} \left( I(r) - \mu_p \right)^2 },
\label{eq:homogeneity}
\end{equation}

\noindent where $\mu_p$ be the mean intensity in the neighborhood, formulated as: 
\[\mu_p = \frac{1}{|N|} \sum_{r \in N_{11 \times 11}(p)} I(r)\]

\noindent The total cost $C_{t}$ is then defined as a weighted sum of equations~\eqref{eq:location}, \eqref{eq:intensity}, and \eqref{eq:homogeneity} normalized over $P$:

\begin{equation}
C_{t}(p) = \alpha \cdot \hat{C}_{l}(p) + \beta \cdot \hat{C}_{i}(p) + \gamma \cdot \hat{C}_{h}(p),
\label{eq:total}
\end{equation}

\noindent where:
\begin{itemize}
    \item $\hat{C}_{l}(p)$, $\hat{C}_{i}(p)$, and $\hat{C}_{h}(p)$ are the min-max normalized values of $C_{l}(p)$, $C_{i}(p)$, and $C_{h}(p)$ over $P$:
    \[
    \hat{C}_{l,i,h}(p) = \frac{C_{l,i,h}(p) - \min\limits_{p' \in P} C_{l,i,h}(p')}{\max\limits_{p' \in P} C_{l,i,h}(p') - \min\limits_{p' \in P} C_{l,i,h}(p')}
    \]
    
    \item $\alpha$, $\beta$, and $\gamma$ are empirically determined weights, set to $\alpha = 1$, $\beta = 1$, and $\gamma = 2$.
\end{itemize}

\noindent For the selection of negative points, we exclude background points with intensities lower than an adaptive threshold $T_N$ that excludes the lowest 10\% of intensities, defined as:
\[T_N = \min(I) + 0.1 \cdot \left( \max(I) - \min(I) \right)\]

\noindent Finally, the optimal prompt location $p^*$ is selected as the point that minimizes equation~\eqref{eq:total}, as follows:

\[p^* = \arg\min_{p \in P} C_{t}(p)\]

\noindent A visual overview of SAMONAI is shown in figure~\ref{fig:samonai}.

\subsection{Radiomics Pipeline}

\begin{figure*}[tb]
    \centering
    \includegraphics[width=\textwidth]{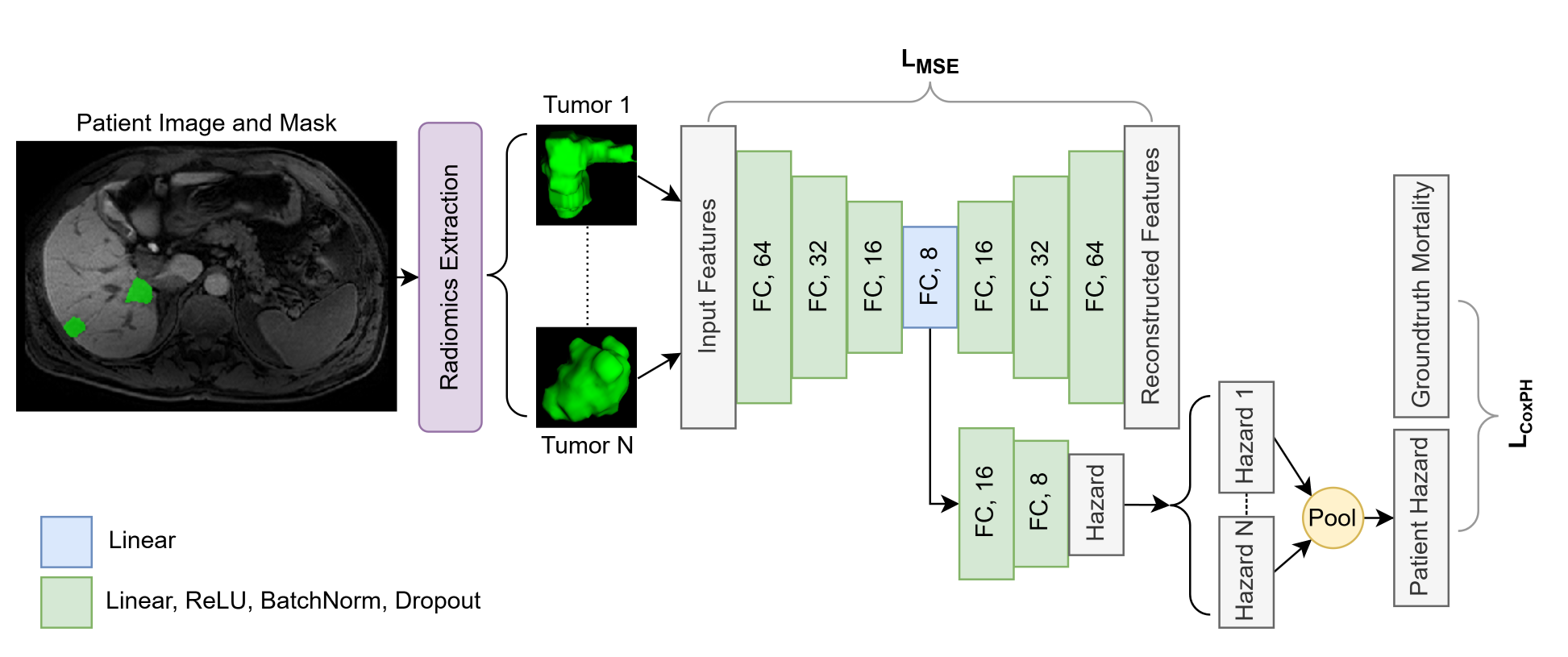}
    \caption{Radiomics pipeline. Radiomic features are extracted independently from each tumor. An autoencoder is used to compress features and a hazard regressor connected to the autoencoder's bottleneck is used to predict hazard from each tumor. Patient-level hazard is calculated by pooling tumor-level hazards. The autoencoder and regressor are jointly trained using a weighted sum of mean squared error and cox proportional hazard loss functions.}
    \label{fig:radiomics_pipeline}
\end{figure*}

To predict post-surgical survival from pre-surgical MRI, we design a radiomics pipeline based on the available segmentations. Our pipeline extracts radiomic features from each tumor in pre- and post-contrast MRI, then trains an autoencoder-based multiple instance network to jointly learn dimensionality reduction and survival prediction for unifocal and multifocal CRLM. An overview of our pipeline is shown in figure~\ref{fig:radiomics_pipeline}

\subsubsection{Radiomics Extraction}
We extract radiomic features from each tumor using PyRadiomics (v3.1.0) \citep{VanGriethuysen2017PyRadiomics}. When extracting features from predicted segmentations, very small predictions—specifically those with a longest diameter $\leq$ the $1^{st}$ percentile of the ground truth segmentations (computed from the training set)—are excluded. This step is essential, as tumor segmentation algorithms often produce numerous small false positives. Radiomic feature extraction on such small regions is problematic because these regions often lack sufficient voxel counts, leading to unreliable or undefined feature values. Additionally, texture and shape features become unstable or meaningless when computed on such limited data, as they are sensitive to noise, interpolation artifacts, and quantization errors. As a result, features extracted from these tiny false positives are often not biologically or diagnostically informative.

For radiomics extraction, the 3D volumes were resampled to an isotropic spacing of $2\,\text{mm}$ using B-spline interpolation to ensure spatial consistency. Voxels with intensities outside the range $[\mu - 3\sigma, \mu + 3\sigma]$, where $\mu$ and $\sigma$ are the mean and standard deviation within the region of interest, were excluded to reduce the influence of outliers. Images were normalized using z-score normalization and subsequently scaled by a factor of 100 to facilitate consistent discretization. A fixed bin width of 5 was used for gray-level discretization. The following feature classes were extracted from the original image: first-order statistics, shape-based, gray level co-occurrence matrix (GLCM), gray level run length matrix (GLRLM), gray level size zone matrix (GLSZM), and gray level dependence matrix (GLDM). Table~\ref{tab:radiomics_summary} summarizes the number of features for each feature class.

\begin{table}[ht]
\centering
\caption{Summary of extracted radiomic features.}
\resizebox{0.9\linewidth}{!}{%
\begin{tabular}{|c|c|c|}
\hline
\textbf{Feature Class} & \textbf{Category} & \textbf{Number of Features} \\
\hline
First-order      & Statistical & 18 \\
Shape-based & Shape       & 14 \\
GLCM             & Textural    & 22 \\
GLRLM            & Textural    & 16 \\
GLSZM            & Textural    & 16 \\
GLDM             & Textural    & 14 \\
\hline
\multicolumn{2}{|c|}{\textbf{Total}} & \textbf{100} \\
\hline
\end{tabular}%
}
\label{tab:radiomics_summary}
\end{table}

Since radiomic features tend to be right-skewed, applying a logarithmic transformation is a common practice \citep{Chen2021AMINN, Kirienko2021LogTransform}. Therefore, we employ a two-step normalization strategy following \citet{Chen2021AMINN}: first, we shift each feature to ensure positivity and then apply a logarithmic transformation; second, we perform z-score normalization. The process is defined as follows:

\noindent Let $\mathbf{x} \in \mathbb{R}^{n \times d}$ denote the feature matrix, where $n$ is the number of samples and $d$ is the number of features. We define the logarithmic transformation as:
\[
\tilde{\mathbf{x_i}} = \log\left(\max(\mathbf{x_i} - \min\mathbf{x_i} + \operatorname{median}(\mathbf{x_i} - \min\mathbf{x_i}), \varepsilon)\right)
\]

\noindent Afterwards, two-step normalized features $\hat{\mathbf{x_i}}$ are calculated by applying z-score normalization over the logarithm transformed features $\tilde{\mathbf{x_i}}$ as:
\[
\hat{\mathbf{x_i}} = \frac{\tilde{\mathbf{x_i}} - \boldsymbol{\mu_i}}{\boldsymbol{\sigma_i}},
\]

\noindent where:
\begin{itemize}
    \item $i \in d$ is the index of a given feature.
    \item $\varepsilon = 10^{-6}$ is a small constant used to clip values $\leq 0$ for numerical stability.
    \item $\boldsymbol{\mu_i}$ and $\boldsymbol{\sigma_i}$ are the mean and standard deviation of the log transformed feature $\tilde{\mathbf{x_i}}$
\end{itemize}

\noindent This normalization is consistently applied to validation and test sets using statistics derived from the training set, avoiding data leakage.

\subsubsection{Outcome Prediction}

Typically, outcome prediction via radiomics involves applying a dimensionality reduction or a feature selection method, such as principal component analysis (PCA) or K-best selection, followed by a machine learning model, such as support vector machines (SVM) or random forests. However, independent feature selection and modeling can lead to discarding critical features for predicting survival. Therefore, we design SurvAMINN, a fully-connected neural network that consists of an autoencoder, for dimensionality reduction, and a MIL regressor, for survival prediction from multifocal CRLM. SurvAMINN extends AMINN~\citep{Chen2021AMINN} to model censored survival data and estimate patient-level hazards by focusing on the most aggressive tumors.

The autoencoder is trained to compress input features into a latent representation and then reconstruct them. Its objective is to minimize the mean squared error loss $\mathcal{L}_{\text{\tiny MSE}}$, defined as:

\begin{equation}
\mathcal{L}_{\text{\tiny MSE}} = \frac{1}{N} \sum_{i=1}^{N} | x_i - \hat{x}_i |^2,
\label{eq:MSE}
\end{equation}

\noindent where $x_i$ and $\hat{x}_i$ are the original and reconstructed feature vectors, respectively, and $N$ is the number of samples.

The multiple instance learning (MIL) regressor is trained to predict tumor-level hazards from the learned latent representations of different tumors. Afterwards, these tumor-level predictions are pooled to compute a patient-level hazard score. 

Since the goal is to identify patients' hazards based on the most lethal tumors, we define the final patient-level hazard as a soft approximation of the maximum tumor-level hazard per patient. Specifically, we employ the \textit{LogSumExp} (LSE) pooling operation to balance robustness and training stability while accounting for patients with multiple high-risk tumors. The patient-level hazard $\eta_p$ is computed from tumor-level hazards $\{\eta_1, \eta_2, \ldots, \eta_n\}$ as follows:

\begin{equation}
\eta_p = \log\left( \sum_{i=1}^{n} \exp(\eta_i) \right),
\label{eq:LSE}
\end{equation}

\noindent This formulation allows the model to focus on the worst tumors while preserving differentiability for optimization. 

The regressor is optimized using a Cox proportional hazard loss $\mathcal{L}_{\text{\tiny CoxPH}}$, defined as:

\begin{equation}
\mathcal{L}_{\text{\tiny CoxPH}} = - \sum_{i: \delta_i = 1} \left( \eta_i - \log \sum_{j \in \mathcal{R}(T_i)} e^{\eta_j} \right),
\label{eq:cox}
\end{equation}

\noindent where:
\begin{itemize}
    \item $\eta_i$ is the predicted hazard for patient $i$.
    \item $\delta_i$ is the mortality status (1 if mortality occurred, 0 if censored).
    \item $T_i$ is the observed time.
    \item $\mathcal{R}(T_i)$ is the hazards of patients still at risk at $T_i$.
\end{itemize}

\noindent The overall objective function $\mathcal{L}$ is a weighted combination of the two losses defined in equations~\eqref{eq:MSE} and \eqref{eq:cox}, formulated as:

\[
\mathcal{L} = (1 - \alpha) \cdot \mathcal{L}_{\text{\tiny MSE}} + \alpha \cdot \mathcal{L}_{\text{\tiny CoxPH}},
\]

\noindent where $\alpha = \frac{\text{current epoch}}{\text{total epochs}} \in [0, 1]$ is a scheduling parameter that gradually shifts the focus from reconstruction to survival prediction, improving training stability.

\subsection{Experimental Setup}

\subsubsection{Evaluation Metrics}

To evaluate the segmentation quality, we compute the Dice-Sørensen Coefficient (dice score) separately for each anatomical structure of interest (e.g., liver, tumor, spleen). The dice score for a given class $c$ is defined as:

\[
\text{Dice}_c = \frac{2|\mathbf{P}_c \cap \mathbf{G}_c|}{|\mathbf{P}_c| + |\mathbf{G}_c|},
\]
\noindent where $\mathbf{P}_c$ denotes the set of predicted voxels belonging to class $c$, and $\mathbf{G}_c$ denotes the set of ground-truth voxels for the same class.

Beyond segmentation accuracy, we evaluate tumor detection performance by treating individual tumor instances as detection targets. A predicted tumor is considered a true positive (TP) if it achieves a dice score $\geq0.1$ with a ground truth tumor. To enforce one-to-one matching, each ground truth tumor is paired with at most one predicted tumor — the one with the highest dice score above the 0.1 threshold. Any remaining unmatched predicted tumors are counted as false positives (FP), and unmatched ground truth tumors are counted as false negatives (FN).

Based on this pairing strategy, we compute standard classification metrics for detection evaluation: precision, recall, and F1-score, as defined below:

\[
\text{Precision} = \frac{\text{TP}}{\text{TP} + \text{FP}} \quad
\text{Recall} = \frac{\text{TP}}{\text{TP} + \text{FN}}
\]
\[
\text{F1-score} = \frac{2 \cdot \text{Precision} \cdot \text{Recall}}{\text{Precision} + \text{Recall}}
\]

For outcome prediction, we use the concordance index (C-index) to assess the predictive performance of our method. The C-index, defined as the proportion of all usable patient pairs whose predicted risks are correctly ordered, is computed as:

\[
\text{C-index} = \frac{\sum_{i,j} \mathbbm{1}_{T_j < T_i} \cdot \mathbbm{1}_{\eta_j > \eta_i} \cdot \delta_j}{\sum_{i,j} \mathbbm{1}_{T_j < T_i} \cdot \delta_j},
\]

\noindent where:
\begin{itemize}
  \item $T_{i,j}$ is the observed survival time for patients$_{i,j}$  
  \item $\eta_{i,j}$ is the predicted hazard for patients$_{i,j}$
  \item $\delta_j$ is the mortality indicator (1 if mortality occurred, 0 if censored) for patient$_j$
  \item $\mathbbm{1}_{\text{condition}}$ equals 1 if the condition is true and 0 otherwise
\end{itemize}

\subsubsection{Training Details}

\noindent \textbf{Segmentation.} Due to the high computational cost, all segmentation models were trained once and validated using a randomly sampled 25\% hold-out validation set, as illustrated in figure~\ref{fig:dataset}~(b). The checkpoint that achieved the highest validation dice score was selected for final evaluation on the test set. All models were trained for 1000 epochs with a batch size of 2. Validation was performed every 2 epochs. Optimization was carried out using the AdamW optimizer with a learning rate of $1 \times 10^{-4}$ and a weight decay of $1 \times 10^{-2}$. A cosine annealing learning rate scheduler was employed, with a linear warm-up phase over the first 10\% of the total iterations.

For preprocessing, all volumes were resampled to an isotropic voxel spacing of 1\,mm. Intensity outliers were removed via 99.95th percentile clipping, and intensities were normalized to the range $[-1, 1]$. To reduce computational burden, background regions were cropped. Data augmentation included 3D flipping, rotations, intensity scaling, and shifting. A sliding window inference technique with a window size of $112 \times 112 \times 112$ was used. The only applied post-processing was the removal of any extra-hepatic predictions as well as the removal of very small predicted tumors with overall volume~$< 100~mm^3$. This threshold was empirically assigned and is much smaller than the smallest tumor in the training data. Training was conducted on a single NVIDIA A100 GPU. 

\noindent \textbf{Outcome Prediction.} For survival analysis, experiments were repeated 15 times using different 3-fold cross-validation splits. Given that the pre- and post-contrast tumor segmentations were unpaired, a late fusion strategy was used by aggregating the predicted hazards from independent pre- and post-contrast models. Each model was trained for 250 epochs using AdamW optimizer with a learning rate of $4 \times 10^{-4}$ and weight decay of $1 \times 10^{-3}$. To mitigate class imbalance, each epoch was balanced by sampling an equal number of censored and uncensored data points. A dropout rate of 20\% was used during training. No GPUs were needed to train survival prediction models.

\subsubsection{Statistical Analysis}

To examine the strength of association between individual variables and survival outcomes, we fit Cox Proportional Hazards models \citep{Cox1972Cox} on z-score standardized variables and compare their Hazard Ratios (HR).

We assess the statistical significance of our findings through multiple approaches. First, we use the Wilcoxon rank-sum test \citep{Mann1947WilcoxonRankSum}, a non-parametric unpaired test, to compare performance between different input settings (i.e., pre/post-contrast vs. both contrasts). Second, we employ a randomization test by training and evaluating our model $1 \times 10^{3}$ different times on data with randomly shuffled survival labels to establish a null distribution. Similar to our actual setup, each random C-index was calculated by averaging scores from 15 different 3-fold cross-validation splits. 

We further stratify patients into high and low-risk groups by median dichotomizing the hazards predicted by SurvAMINN, averaged over the 15 runs. The groups are then compared using Kaplan–Meier curves and the log-rank test. Finally, to evaluate the robustness of HR estimates across repeated runs, we compute 95\% confidence intervals (95\% CI) of $1 \times 10^{3}$ bootstrapped HR estimates, and the two-sided \emph{P}-values are calculated using the Wald test, as follows:

\[
z = \frac{\log(\hat{\text{HR}})}{\text{SD}(\log(\text{HR}))},
\quad
p = 2 \cdot \left(1 - \Phi\left(|z|\right)\right)
\]

\noindent where $\hat{\text{HR}}$ is the mean of the bootstrapped hazard ratios, $\text{SD}(\log(\text{HR}))$ is the sample standard deviation of the log-transformed HR values, and $\Phi$ denotes the cumulative distribution function of the standard normal distribution.

\section{Results}
\label{sec:results}

\subsection{Segmentation}

\begin{figure*}[tb]
    \centering
    \includegraphics[width=\textwidth]{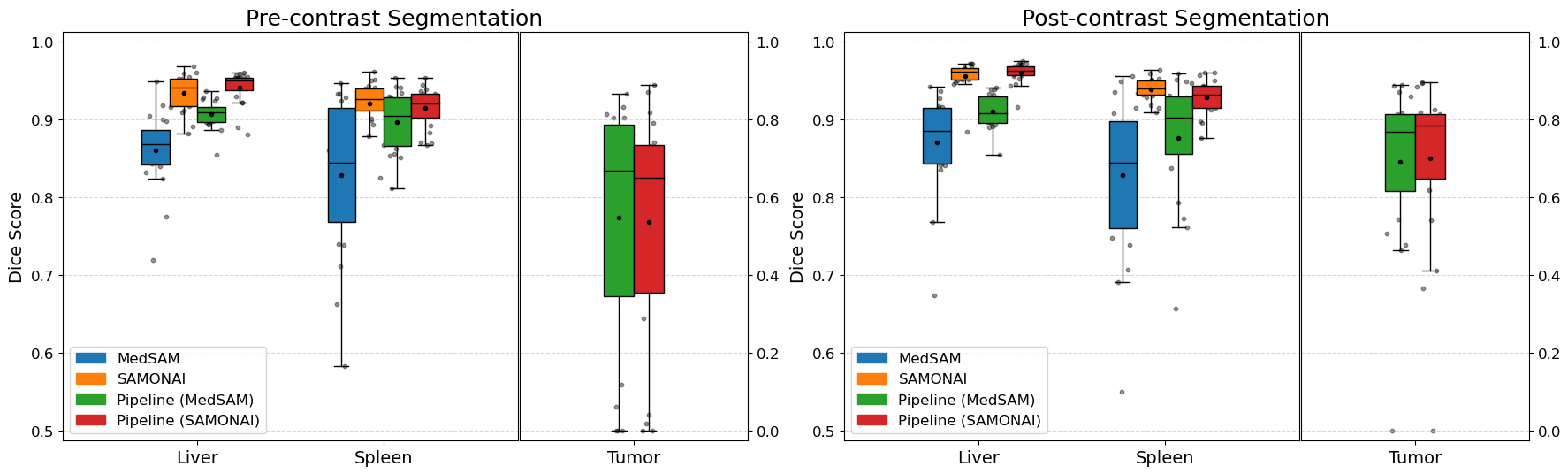}
    \caption{Box plot of the segmentation dice scores. Grey dots represent individual cases, and black dots indicate their mean values.}
    \label{fig:segmentation_results}
\end{figure*}

Segmentation algorithms were evaluated using the 20 fully annotated cases, indicated as (1) and (2) in figure~\ref{fig:dataset}. Due to the high cost of manual annotation, liver and spleen labels were available only for the post-contrast images. To enable evaluation of pre-contrast segmentation algorithms, we propagated these labels to the pre-contrast images via image registration. A two-step registration process—affine followed by deformable—was performed using Elastix \citep{Klein2010Elastix}.

Figure~\ref{fig:segmentation_results} presents the dice scores for liver, tumor, and spleen segmentation using our pipeline on pre- and post-contrast MRI images. We also compare the performance of SAMONAI and MedSAM on missing labels (i.e., liver and spleen). SAMONAI outperforms MedSAM with mean dice score improvements of over 7\% for liver and 9\% for spleen on pre-contrast images, and 9\% for liver and 11\% for spleen on post-contrast images. This improvement drives the strong performance of our pipeline, which achieves mean dice scores of 0.94 for liver and 0.92 for spleen on pre-contrast images, and 0.96 for liver and 0.93 for spleen on post-contrast images. 

Using SAMONAI in stage 1, compared to using MedSAM, improves the pipeline's mean dice scores for both pre- and post-contrast MRI. For pre-contrast segmentation, dice scores increase by 8\% for the liver and 9\% for the spleen. For post-contrast MRI, improvements of 5\% are observed for both the liver and the spleen. Tumor segmentation also yields satisfactory results, with mean dice scores of 0.54 for pre-contrast and 0.70 for post-contrast images. Since SAMONAI contributes only to missing labels, training the pipeline with either MedSAM or SAMONAI results in similar tumor segmentation performance.

Moreover, we report the tumor detection performance of our segmentation pipeline, as shown in table~\ref{tab:detection_results}. As expected, the post-contrast results surpass those of the pre-contrast images, with an F1-score of 0.79, due to the enhanced tumor visibility. Nevertheless, the detection performance on pre-contrast images, with an F1-score of 0.71, remains satisfactory despite the inherent challenges in distinguishing tumors without contrast enhancement. These results highlight the potential of pre-contrast predictions to contribute meaningfully to outcome prediction.

\begin{table}[ht]
\centering
\resizebox{\linewidth}{!}{
\begin{tabular}{lcccccc}
\hline
\textbf{Contrast} & \textbf{Precision} & \textbf{Recall} & \textbf{F1-score} & \textbf{TP} & \textbf{FP} & \textbf{FN} \\
\hline
Pre-contrast  & 0.674 & 0.756 & 0.713 & 31 & 15 & 10 \\
Post-contrast & 0.720 & 0.878 & 0.791 & 36 & 14 & 5 \\
\hline
\end{tabular}}
\caption{Tumor detection results of the segmentation pipeline. TP, FP, and FN refer to the overall number of true positives, false positives, and false negatives across the test set.}
\label{tab:detection_results}
\end{table}

\begin{figure*}[tb]
    \centering
    \includegraphics[width=\textwidth]{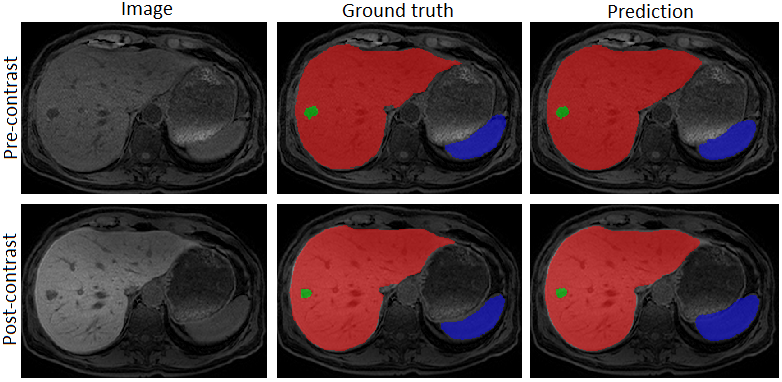}
    \caption{Qualitative assessment of the segmentation pipeline on a sample testing case.}
    \label{fig:qualitative}
\end{figure*}

We also provide a visual comparison of the predicted segmentations against the ground truth for a randomly selected test case, shown in figure~\ref{fig:qualitative}. The dice scores for the pre-contrast segmentation are 0.956 for liver, 0.742 for tumors, and 0.915 for spleen, while the post-contrast scores are 0.963, 0.786, and 0.922, respectively. The precise delineation observed in both contrast settings highlights the robustness of our pipeline and its potential to rival fully supervised approaches.

\subsection{Outcome Prediction}

In evaluating our outcome prediction pipeline, unless explicitly mentioned, we utilize predicted segmentations from both pre- and post-contrast MRI. We evaluate our hypotheses through four sets of experiments. 

\begin{figure}[htb]
    \centering
    \includegraphics[width=\columnwidth]{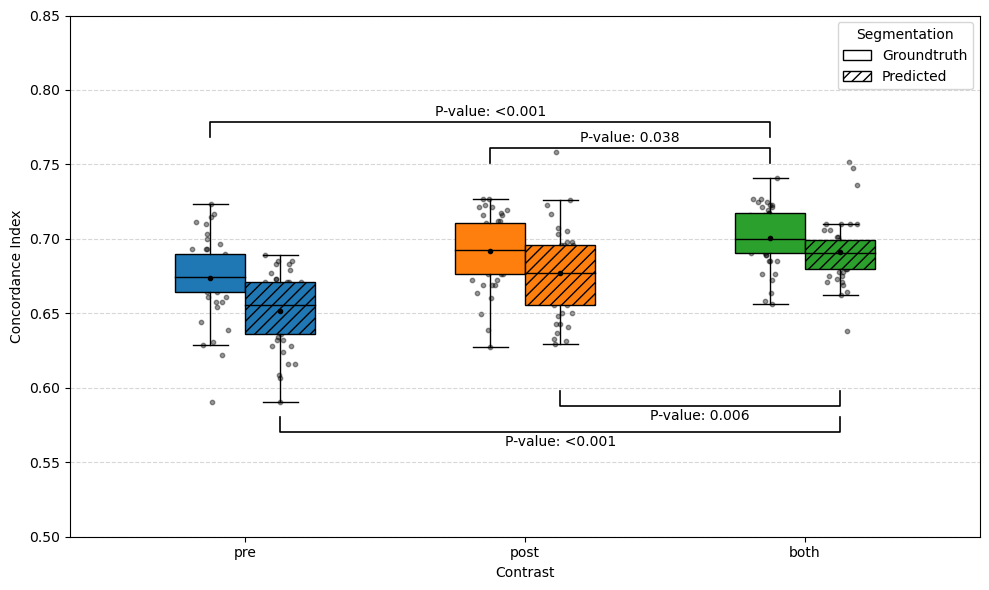}
    \caption{Box plot of outcome prediction C-index for pre-contrast (pre), post-contrast (post), and combined contrast phases (both). Grey dots represent individual runs, and black dots indicate their mean values. Striped boxes correspond to results obtained using predicted segmentations. Statistical significance, assessed using the Wilcoxon rank-sum test's p-value, confirms that combining pre- and post-contrast MRI yields better performance than using either phase alone.}
    \label{fig:radiomics_results}
\end{figure}

\begin{figure*}[tb]
    \centering
    \includegraphics[width=\textwidth]{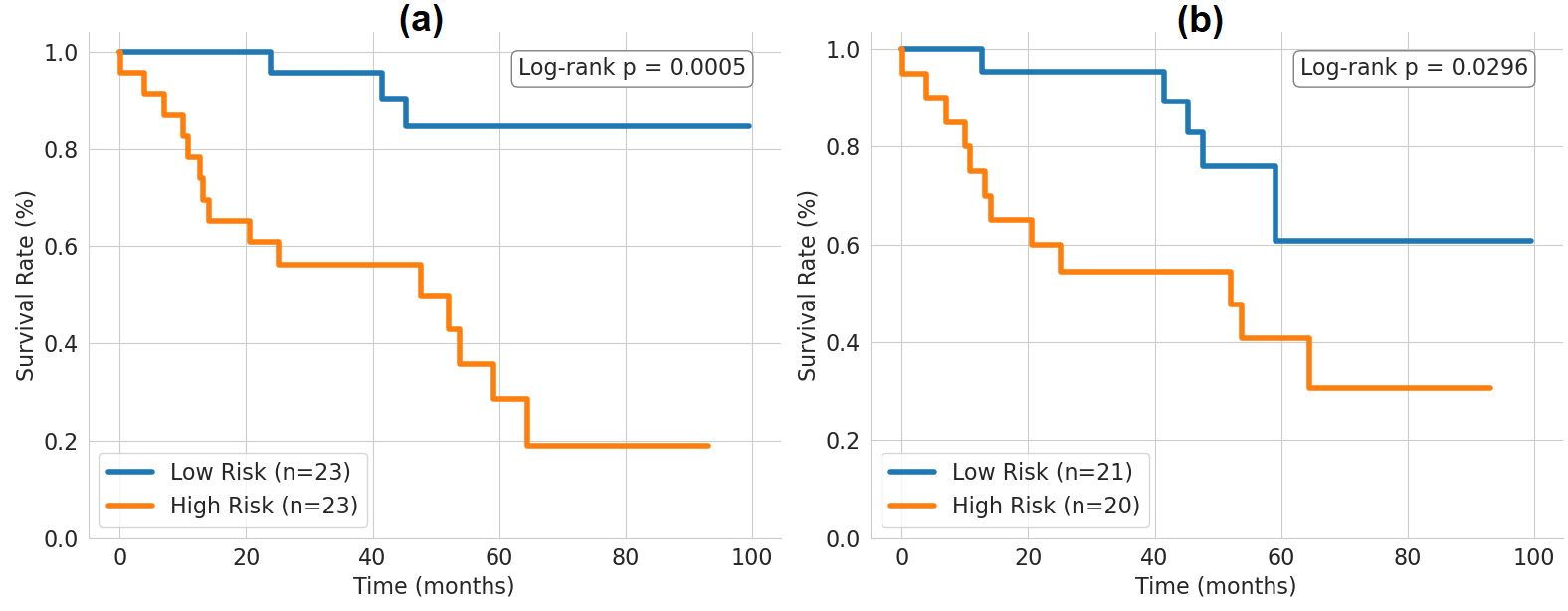}
    \caption{Kaplan–Meier curves of our outcome prediction pipeline using (a) ground truth segmentations and (b) predicted segmentations. Five patients, for whom no tumors were predicted (i.e., empty segmentations), were excluded from the analysis in (b), as hazard scores could not be computed.}
    \label{fig:kaplan_meier}
\end{figure*}

First, we compare outcome prediction using pre- and post-contrast phases individually versus jointly. Figure~\ref{fig:radiomics_results} presents the results of our pipeline using both ground truth and predicted segmentations. As expected, post-contrast segmentations yield better outcome prediction than pre-contrast alone; however, combining both phases consistently improves performance, as confirmed by statistical significance tests. Although replacing ground truth with predicted segmentations leads to a slight drop in performance—0.023 for pre-contrast, 0.015 for post-contrast, and 0.010 when using both—the pipeline retains strong predictive capability. Figure~\ref{fig:kaplan_meier} reflects our pipeline's capabilities to stratify patients into low and high risk groups.

Second, we compare our tumor-level to patient-level hazard pooling strategy to other alternatives. Table~\ref{tab:pooling_results} shows the results when predicting patient-level hazard based on the largest tumor, mean of tumor hazards, and max tumor hazard. Emphasizing worst tumors, according to the model, with max pooling outperforms using the largest tumor or the average of all of the tumors. LSE pooling further improves the results by considering multiple bad tumors, improving stability and accuracy.

\begin{table}[htb]
\centering
\resizebox{0.9\linewidth}{!}{
\begin{tabular}{lccc}
\hline
\textbf{Pooling} & \textbf{C-Index} & \textbf{HR (95\% CI)} & \textbf{P-val} \\
\hline
mean        & 0.556 & 1.114 (1.040--1.216) & 0.759   \\
largest     & 0.594 & 1.263 (1.147--1.391) & 0.618   \\
max         & 0.623 & 1.568 (1.441--1.784) & 0.162   \\
LSE   & 0.691 & 1.929 (1.890--1.980) & $<$0.001 \\
\hline
\end{tabular}
}
\caption{Comparison of different pooling methods.}
\label{tab:pooling_results}
\end{table}

Third, as a baseline, we assess outcome prediction performance using independent dimensionality reduction and modeling steps. For feature reduction, we evaluate four approaches: no reduction (all features), K-best selection, PCA, and minimal-redundancy-maximal-relevance (mRMR) \citep{Peng2005mRMR}. Each feature set is used to train two models: SVM for 3-year survival classification, and a random survival forest (RSF) model \citep{Ishwaran2008RandomSurvivalForests}, which extends random forests to handle censored survival data. Table~\ref{tab:baselines} compares these baselines to our proposed SurvAMINN. While the SVM generally outperforms RSF, its performance is less stable, likely due to the reduced sample size resulting from censoring. In contrast, SurvAMINN jointly learns both dimensionality reduction and survival prediction using the full dataset, enabling it to consistently outperform all baseline approaches.

\begin{table}[ht]
\centering
\resizebox{\linewidth}{!}{
\begin{tabular}{ccccc}
\hline
\textbf{Model} & \textbf{Reduction} & \textbf{C-Index} & \textbf{HR (95\% CI)} & \textbf{P-val} \\
\hline
SVM  & None     & 0.659 & 1.781 (1.693--1.876) & 0.002     \\
SVM  & KBest    & 0.647 & 1.745 (1.652--1.842) & 0.005     \\
SVM  & PCA      & 0.667 & 1.907 (1.793--2.036) & 0.004     \\
SVM  & mRMR     & 0.634 & 1.568 (1.494--1.652) & 0.011     \\
RSF  & None     & 0.632 & 1.447 (1.368--1.531) & 0.072     \\
RSF  & KBest    & 0.637 & 1.543 (1.453--1.653) & 0.056     \\
RSF  & PCA      & 0.634 & 1.475 (1.412--1.540) & 0.012     \\
RSF  & mRMR     & 0.631 & 1.446 (1.373--1.523) & 0.052     \\
\multicolumn{2}{c}{\textbf{SurvAMINN}} & \textbf{0.691} & \textbf{1.929 (1.890--1.980)} & \textbf{$<$0.001} \\

\hline
\end{tabular}
}
\caption{Comparison of SurvAMINN against common methods. None refers to using the whole set of features without dimensionality reduction.}
\label{tab:baselines}
\end{table}

Fourth, we compare our predicted method to existing clinical and molecular biomarkers, as shown in table~\ref{tab:biomarkers}. Although gene mutations, such as APC, TP53, KRAS, and NRAS, are acquired after surgery from resected samples, they fail to accurately predict survival. While the overall predictive performance, measured by C-index, improves when combining SurvAMINN with these biomarkers in a multivariate Cox model, SurvAMINN remains the only predictor with strong and statistically significant association with outcome, as revealed by the hazard ratios and p-values.

\begin{table*}[tb]
\centering
\scriptsize
\caption{Comparison of SurvAMINN with existing clinical and genomic biomarkers using univariate and multivariate Cox models.}
\resizebox{0.8\textwidth}{!}{
\begin{tabular}{lcccccl}
\toprule
\multirow{2}{*}{\textbf{Biomarker}} & \multicolumn{3}{c}{\textbf{Univariate}} & \multicolumn{3}{c}{\textbf{Multivariate}} \\
\cmidrule(lr){2-4} \cmidrule(lr){5-7}
 & C-Index & HR (95\% CI) & P-val & C-Index & HR (95\% CI) & P-val \\
\midrule
Sex              & 0.476 & 0.93 (0.57--1.51) & 0.768 & \multirow{8}{*}{\textbf{0.751}} & 1.16 (1.12--1.20) & 0.691 \\
Age              & 0.525 & 1.34 (0.81--2.22) & 0.258 & & 1.54 (1.50--1.58) & 0.164 \\
Fong             & 0.545 & 1.17 (0.74--1.85) & 0.500 & & 0.91 (0.88--0.94) & 0.741 \\
TTE              & 0.538 & 1.21 (0.81--1.80) & 0.347 & & 1.12 (1.09--1.17) & 0.737 \\
APC              & 0.583 & 0.82 (0.50--1.34) & 0.418 & & 0.59 (0.58--0.61) & 0.127 \\
TP53             & 0.502 & 1.28 (0.80--2.07) & 0.306 & & 1.71 (1.65--1.85) & 0.163 \\
KRAS             & 0.538 & 0.93 (0.57--1.54) & 0.785 & & 0.88 (0.86--0.90) & 0.668 \\
NRAS             & 0.500 & 1.17 (0.85--1.60) & 0.339 & & 0.85 (0.82--0.88) & 0.634 \\
\textbf{SurvAMINN} & \textbf{0.691} & \textbf{1.93 (1.89--1.98)} & \textbf{$<$0.001} & & \textbf{3.41 (3.17--4.02)} & \textbf{$<$0.001} \\
\bottomrule
\end{tabular}
}
\label{tab:biomarkers}
\end{table*}

\begin{figure}[htb]
    \centering
    \includegraphics[width=\columnwidth]{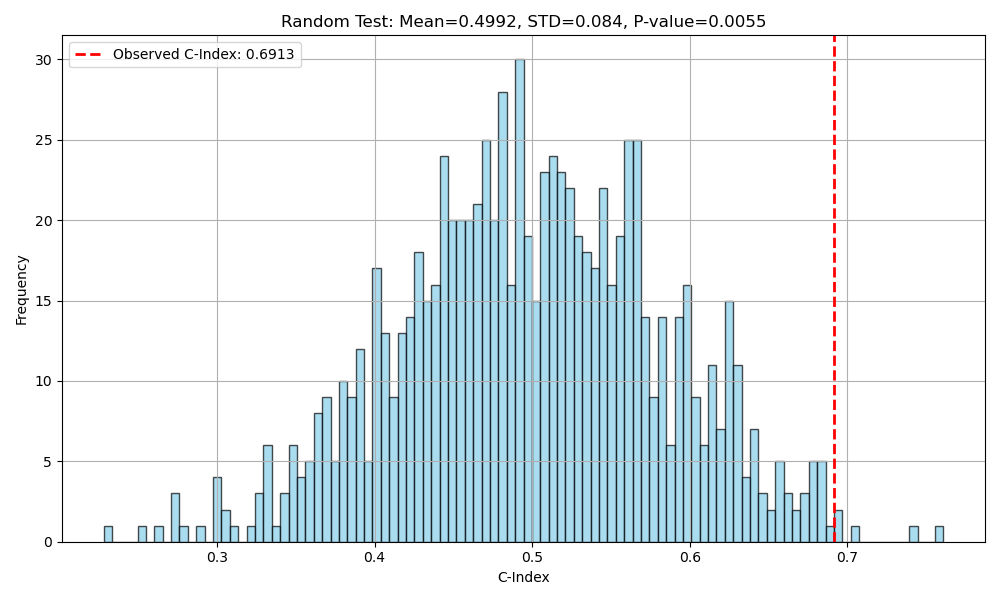}
    \caption{Randomized testing of SurvAMINN. Blue bars represent the C-index distribution when training and testing on randomly shuffled labels. Red line represents the observed score without randomization.}
    \label{fig:random_test}
\end{figure}

Last but not least, we perform a randomized test of SurvAMINN, as shown in figure~\ref{fig:random_test}. The test further confirms the statistical significance of the achieved C-index.

\section{Discussion}
\label{sec:discussion}

This study presents a fully-automated framework to predict surgical outcomes of CRLM from pre-surgical MRI. Our results reveal that developing an anatomy-aware segmentation method is essential to automate survival analysis with minimal sacrifice in predictive capability.

On one hand, our zero-shot 3D prompt propagation method, SAMONAI, achieved substantial accuracy in segmenting the liver and the spleen. Remarkably, SAMONAI was able to outperform medical imaging finetuned models, such as MedSAM, improving accuracy and generalizability. Moreover, incorportating SAMONAI into our partial labels segmentation pipeline yields a fully automated segmentation model for liver, tumor, and spleen segmentation. The accuracy of the liver and spleen segmentations, as shown in figure~\ref{fig:qualitative}, enables accurate elimination of extra-hepatic tumors. This contributes to the satisfactory detection performance reported in table~\ref{tab:detection_results}. 

In addition to the evaluation of tumor detection described in table~\ref{tab:detection_results}, the post-contrast predicted segmentations were manually assessed by the same radiologist with 13 years of abdominal imaging experience who performed the manual segmentations described in section~\ref{subsec:dataset}. The radiologist's assessment revealed significantly higher detection performance than our estimations. Our detection metrics count all disconnected objects as independent tumors. Therefore, in case of noisy or ill-defined segmentation borders, this strategy overestimates FPs and FNs by counting random disconnected pixels as tumors. The expert's assessment revealed that 13 FPs and 6 FNs were not real but rather single pixels or disconnected borders.

Moreover, another source of FPs is that CRLMs with longest diameter $\leq 1~cm$ are not annotated in the ground truth segmentations. Our pipeline detected 7 small CRLM tumors that were counted as FPs. Furthermore, benign lesions are not included in the manual segmentations. Our pipeline segmented 6 benign lesions (3 cysts and 3 hemangiomas) leading to 6 additional FPs. While benign lesions minimally contribute to prognosis, they are not expected to affect our surgical outcome pipeline, especially since our pipeline estimates patient-level hazards based on the most lethal tumors, as described in equation~\eqref{eq:LSE}.

Although the small object removal step reduced FPs from 29 to 14 and FNs from 8 to 5, many of the remaining errors were still attributable to the factors discussed above, primarily due to the conservative thresholds used for eliminating small objects. In fact, the manual assessment revealed that only 3 FPs and 2 FNs were truly false, after excluding the misclassified ones according to the review. This suggests that the actual precision and recall are \textasciitilde0.93 and \textasciitilde0.95, respectively. 

Interestingly, the radiologist's analysis revealed that 2 of the actual FPs corresponded to surgical resection margins in a patient who had undergone prior hepatectomy. Notably, liver segmentation for this case was highly accurate (dice score $\approx 0.97$). The third FP, observed in a different case, was attributed to biliary duct dilation. These 3 were the only truly incorrect FPs, likely due to the absence of similar abnormalities in the training data.

These findings highlight the robustness of our segmentation pipeline, even in challenging clinical scenarios. By eliminating the need for labor-intensive manual segmentation, reducing reliance on expert input, and minimizing inter-rater variability, our approach streamlines clinical workflows. Furthermore, the accuracy and precision of our pipeline form the foundation for developing a fully automated method for surgical outcome prediction.

On the other hand, the results of the outcome prediction pipeline also confirm our four hypotheses. For instance, utilizing both pre- and post-contrast phases results in a statistically significant improvement over using a single one, as shown in figure~\ref{fig:radiomics_results}. Moreover, worst tumors, according to our hazard prediction model, SurvAMINN, are more predictive of overall survival than largest tumors or average of all tumors, as shown in table~\ref{tab:pooling_results}. Additionally, the improvement of LSE over max pooling suggests that multiple similarly bad tumors further impair overall survival. 

Furthermore, table~\ref{tab:baselines} highlights that the joint learning of dimensionality reduction and survival prediction outperforms independent feature selection and modeling. This suggests that these independent dimensionality reduction algorithms can potentially lose features that are crucial for outcome prediction. Finally, SurvAMINN proves to be more predictive of outcomes than existing weak clinical and molecular biomarkers shown in table~\ref{tab:biomarkers}.

While our framework was able to automate and improve surgical outcome prediction of CRLM, it can further be improved. For instance, combining pre- and post-contrast images results in a limited improvement. A potential reason could be the limited additional information from pre-contrast images. Since our framework was designed to be generalizable and multimodal, we plan to add more informative modalities, such as T2-weighted MRI and CT, upon availability. Another reason could be the simple late fusion strategy employed due to the pre- and post-contrast misalignment. Therefore, we expect further improvement if an accurate tumor matching algorithm is applied. 

Furthermore, our multivariate survival analysis, as shown in table~\ref{tab:biomarkers}, reveals that radiomic features, along with existing clinical and genomic biomarkers, offer complementary prognostic information. Based on these findings, we plan to integrate clinical and molecular data into our outcome prediction model in future work.

A notable limitation of our framework in calculating relative risks from predicted segmentations is the exclusion of patients for whom the segmentation pipeline does not predict any tumor regions, as they are treated as healthy by the outcome prediction pipeline, as mentioned in figure~\ref{fig:kaplan_meier}. Although this exclusion is necessary for evaluation, it can introduce selection bias and misrepresent the performance of the model. Future work should aim to improve tumor detection sensitivity to reduce such failure cases and enable more comprehensive risk assessment throughout the entire patient cohort.

Our segmentation pipeline segments missing targets by learning from a foundation model. Therefore, inaccuracies in the segmentation generated by the foundation model in stage 1 can impair the final performance. Therefore, in future work, we plan to implement a quality control step, to reduce the influence of failure cases. Moreover, since stage 1 utilizes a randomly sampled subset of the data, employing active learning strategies to select the most informative samples could enhance performance. Quality control can also be beneficial for SAMONAI, since it assumes accurate segmentation of axial, coronal, and sagittal slices in steps 1 and 2 before prompt propagation in step 3, which is also subject to the user's initial prompt.

Another potential improvement for our framework is to utilize the available segmentations of the liver and the spleen in feature extraction, since features extracted from background parenchyma can potentially correlate with surgical morbidity, response to neoadjuvant or adjuvant chemotherapy, and long-term outcome. Therefore, we plan to revisit our radiomics extraction stage in future work.

Finally, we aim to extend our framework to predict response to chemotherapy in non-surgical patient cohorts. This extension can enhance the clinical utility of our approach by supporting treatment decision-making for patients who are not candidates for surgery, ultimately contributing to more personalized and effective care strategies.
\section{Conclusions}
\label{sec:conclusions}

This study describes a fully-automated framework for CRLM surgical outcome prediction from non-contrast and contrast-enhanced MRI. The framework consists of a segmentation pipeline followed by a radiomics pipeline. The segmentation pipeline learns to segment the liver, tumors, and spleen from partial labels by utilizing promptable foundation models to annotate missing targets. Moreover, we propose SAMONAI, a novel zero-shot 3D prompt propagation algorithm that segments 3D regions with a single point prompt. Incorporating SAMONAI into our pipeline further improves its accuracy, efficiency, and generalizability. Our pipeline achieves satisfactory segmentation and detection performance for pre- and post-contrast MRI, making it suitable for radiomics-based outcome prediction. Our radiomics pipeline extracts features from predicted pre- and post-contrast tumor segmentations and predicts survival using SurvAMINN, a neural network that jointly learns dimensionality reduction and outcome prediction from censored data. Furthermore, in case of multifocal CRLM, SurvAMINN predicts patients hazards based on their worst tumors, making it more accurate and interpretable. Finally, we demonstrate that our framework is more predictive of surgical outcome compared to existing clinical and genomic biomarkers. In conclusion, our study reveals the potential of artificial intelligence to improve and automate surgical outcome prediction of CRLM.

\section*{Acknowledgments}

I would like to sincerely thank my supervisors, Dr.~Anne Martel and Dr.~Helen Cheung, for their invaluable guidance, continuous support, and encouragement throughout the course of this project. I would also like to acknowledge Dr.~Jianan Chen, whose pioneering work laid the foundation for this study and who provided valuable insights during my research. I am deeply indebted to Dr.~Hossam El-Rewaidy for his contributions, expertise, and inspiration, all of which were pivotal to the success of my research. I am grateful to all members of the Martel Lab for their insightful discussions and constructive feedback, which have greatly enriched my work. Finally, I extend my heartfelt appreciation to the MAIA consortium. The knowledge, experiences, and connections I have gained through this program are truly lasting and have made a significant impact on my academic and personal growth.

\bibliography{references}

\begin{thebibliography}{58}
\expandafter\ifx\csname natexlab\endcsname\relax\def\natexlab#1{#1}\fi
\providecommand{\url}[1]{\texttt{#1}}
\providecommand{\href}[2]{#2}
\providecommand{\path}[1]{#1}
\providecommand{\DOIprefix}{doi:}
\providecommand{\ArXivprefix}{arXiv:}
\providecommand{\URLprefix}{URL: }
\providecommand{\Pubmedprefix}{pmid:}
\providecommand{\doi}[1]{\href{http://dx.doi.org/#1}{\path{#1}}}
\providecommand{\Pubmed}[1]{\href{pmid:#1}{\path{#1}}}
\providecommand{\bibinfo}[2]{#2}
\ifx\xfnm\relax \def\xfnm[#1]{\unskip,\space#1}\fi
\bibitem[{Bahbahani et~al.(2023)Bahbahani, Khan, Ahmed, Al-Rabiy, Baqer and Al-Terki}]{Bahbahani2023TumorAggression}
\bibinfo{author}{Bahbahani, B.}, \bibinfo{author}{Khan, R.N.}, \bibinfo{author}{Ahmed, I.}, \bibinfo{author}{Al-Rabiy, F.N.}, \bibinfo{author}{Baqer, M.}, \bibinfo{author}{Al-Terki, A.}, \bibinfo{year}{2023}.
\newblock \bibinfo{title}{Rare renaissance: Quadruple synchronous renal cell tumors in one kidney – a case report}.
\newblock \bibinfo{journal}{Urology Case Reports} \bibinfo{volume}{51}, \bibinfo{pages}{102574}.
\newblock \DOIprefix\doi{10.1016/j.eucr.2023.102574}.
\bibitem[{Chen et~al.(2021)Chen, Cheung, Milot and Martel}]{Chen2021AMINN}
\bibinfo{author}{Chen, J.}, \bibinfo{author}{Cheung, H.M.C.}, \bibinfo{author}{Milot, L.}, \bibinfo{author}{Martel, A.L.}, \bibinfo{year}{2021}.
\newblock \bibinfo{title}{Aminn: Autoencoder-based multiple instance neural network improves outcome prediction in multifocal liver metastases} , \bibinfo{pages}{752--761}.
\bibitem[{Chen et~al.(2019)Chen, Milot, Cheung and Martel}]{Chen2019AEGMM}
\bibinfo{author}{Chen, J.}, \bibinfo{author}{Milot, L.}, \bibinfo{author}{Cheung, H.M.C.}, \bibinfo{author}{Martel, A.L.}, \bibinfo{year}{2019}.
\newblock \bibinfo{title}{Unsupervised clustering of quantitative imaging phenotypes using autoencoder and gaussian mixture model} , \bibinfo{pages}{575--582}.
\bibitem[{Chen et~al.(2020)Chen, Kornblith, Norouzi and Hinton}]{Chen2020SimCLR}
\bibinfo{author}{Chen, T.}, \bibinfo{author}{Kornblith, S.}, \bibinfo{author}{Norouzi, M.}, \bibinfo{author}{Hinton, G.}, \bibinfo{year}{2020}.
\newblock \bibinfo{title}{A simple framework for contrastive learning of visual representations} .
\bibitem[{Chen et~al.(2022)Chen, Duan, Wang, He, Lu, Dai and Qiao}]{Chen2022ViTAdapter}
\bibinfo{author}{Chen, Z.}, \bibinfo{author}{Duan, Y.}, \bibinfo{author}{Wang, W.}, \bibinfo{author}{He, J.}, \bibinfo{author}{Lu, T.}, \bibinfo{author}{Dai, J.}, \bibinfo{author}{Qiao, Y.}, \bibinfo{year}{2022}.
\newblock \bibinfo{title}{Vision transformer adapter for dense predictions}.
\newblock \bibinfo{journal}{arXiv preprint arXiv:2205.08534} .
\bibitem[{Chernyak et~al.(2018)Chernyak, Fowler, Kamaya, Kielar, Elsayes, Bashir, Kono, Do, Mitchell, Singal, Tang and Sirlin}]{Chernyak2018LIRADS}
\bibinfo{author}{Chernyak, V.}, \bibinfo{author}{Fowler, K.J.}, \bibinfo{author}{Kamaya, A.}, \bibinfo{author}{Kielar, A.Z.}, \bibinfo{author}{Elsayes, K.M.}, \bibinfo{author}{Bashir, M.R.}, \bibinfo{author}{Kono, Y.}, \bibinfo{author}{Do, R.K.}, \bibinfo{author}{Mitchell, D.G.}, \bibinfo{author}{Singal, A.G.}, \bibinfo{author}{Tang, A.}, \bibinfo{author}{Sirlin, C.B.}, \bibinfo{year}{2018}.
\newblock \bibinfo{title}{Liver imaging reporting and data system (li-rads) version 2018: Imaging of hepatocellular carcinoma in at-risk patients}.
\newblock \bibinfo{journal}{Radiology} \bibinfo{volume}{289}, \bibinfo{pages}{816--830}.
\newblock \DOIprefix\doi{10.1148/radiol.2018181494}.
\bibitem[{Cheung et~al.(2019)Cheung, Karanicolas, Coburn, Seth, Law and Milot}]{Cheung2019TTE}
\bibinfo{author}{Cheung, H.M.C.}, \bibinfo{author}{Karanicolas, P.J.}, \bibinfo{author}{Coburn, N.}, \bibinfo{author}{Seth, V.}, \bibinfo{author}{Law, C.}, \bibinfo{author}{Milot, L.}, \bibinfo{year}{2019}.
\newblock \bibinfo{title}{Delayed tumour enhancement on gadoxetate-enhanced mri is associated with overall survival in patients with colorectal liver metastases}.
\newblock \bibinfo{journal}{European Radiology} \bibinfo{volume}{29}, \bibinfo{pages}{1032--1038}.
\newblock \DOIprefix\doi{10.1007/s00330-018-5618-5}.
\bibitem[{Cox(1972)}]{Cox1972Cox}
\bibinfo{author}{Cox, D.R.}, \bibinfo{year}{1972}.
\newblock \bibinfo{title}{Regression models and life-tables}.
\newblock \bibinfo{journal}{Journal of the Royal Statistical Society: Series B (Methodological)} \bibinfo{volume}{34}, \bibinfo{pages}{187--202}.
\newblock \DOIprefix\doi{10.1111/j.2517-6161.1972.tb00899.x}.
\bibitem[{Dong and Gruenberger(2023)}]{Dong2023Surg}
\bibinfo{author}{Dong, Y.}, \bibinfo{author}{Gruenberger, T.}, \bibinfo{year}{2023}.
\newblock \bibinfo{title}{Surgical management of colorectal liver metastases—a practical clinical approach}.
\newblock \bibinfo{journal}{European Surgery} \bibinfo{volume}{55}, \bibinfo{pages}{94--99}.
\newblock \DOIprefix\doi{10.1007/s10353-023-00796-w}.
\bibitem[{Dosovitskiy et~al.(2021)Dosovitskiy, Beyer, Kolesnikov, Weissenborn, Zhai, Unterthiner, Dehghani, Minderer, Heigold, Gelly, Uszkoreit and Houlsby}]{Dosovitskiy2021ViT}
\bibinfo{author}{Dosovitskiy, A.}, \bibinfo{author}{Beyer, L.}, \bibinfo{author}{Kolesnikov, A.}, \bibinfo{author}{Weissenborn, D.}, \bibinfo{author}{Zhai, X.}, \bibinfo{author}{Unterthiner, T.}, \bibinfo{author}{Dehghani, M.}, \bibinfo{author}{Minderer, M.}, \bibinfo{author}{Heigold, G.}, \bibinfo{author}{Gelly, S.}, \bibinfo{author}{Uszkoreit, J.}, \bibinfo{author}{Houlsby, N.}, \bibinfo{year}{2021}.
\newblock \bibinfo{title}{An image is worth 16x16 words: Transformers for image recognition at scale} .
\bibitem[{Fong et~al.(1999)Fong, Fortner, Sun, Brennan and Blumgart}]{Fong1999Fong}
\bibinfo{author}{Fong, Y.}, \bibinfo{author}{Fortner, J.}, \bibinfo{author}{Sun, R.L.}, \bibinfo{author}{Brennan, M.F.}, \bibinfo{author}{Blumgart, L.H.}, \bibinfo{year}{1999}.
\newblock \bibinfo{title}{Clinical score for predicting recurrence after hepatic resection for metastatic colorectal cancer: Analysis of 1001 consecutive cases}.
\newblock \bibinfo{journal}{Annals of Surgery} \bibinfo{volume}{230}.
\newblock \DOIprefix\doi{10.1097/00000658-199909000-00004}.
\bibitem[{van Griethuysen et~al.(2017)van Griethuysen, Fedorov, Parmar, Hosny, Aucoin, Narayan, Beets-Tan, Fillion-Robin, Pieper and Aerts}]{VanGriethuysen2017PyRadiomics}
\bibinfo{author}{van Griethuysen, J.J.}, \bibinfo{author}{Fedorov, A.}, \bibinfo{author}{Parmar, C.}, \bibinfo{author}{Hosny, A.}, \bibinfo{author}{Aucoin, N.}, \bibinfo{author}{Narayan, V.}, \bibinfo{author}{Beets-Tan, R.G.}, \bibinfo{author}{Fillion-Robin, J.C.}, \bibinfo{author}{Pieper, S.}, \bibinfo{author}{Aerts, H.J.}, \bibinfo{year}{2017}.
\newblock \bibinfo{title}{Computational radiomics system to decode the radiographic phenotype}.
\newblock \bibinfo{journal}{Cancer Research} \bibinfo{volume}{77}, \bibinfo{pages}{e104--e107}.
\newblock \DOIprefix\doi{10.1158/0008-5472.CAN-17-0339}.
\bibitem[{Hatamizadeh et~al.(2022a)Hatamizadeh, Nath, Tang, Yang, Roth and Xu}]{Hatamizadeh2022SwinUNETR}
\bibinfo{author}{Hatamizadeh, A.}, \bibinfo{author}{Nath, V.}, \bibinfo{author}{Tang, Y.}, \bibinfo{author}{Yang, D.}, \bibinfo{author}{Roth, H.R.}, \bibinfo{author}{Xu, D.}, \bibinfo{year}{2022}a.
\newblock \bibinfo{title}{Swin unetr: Swin transformers for semantic segmentation of brain tumors in mri images} , \bibinfo{pages}{272--284}.
\bibitem[{Hatamizadeh et~al.(2022b)Hatamizadeh, Tang, Nath, Yang, Myronenko, Landman, Roth and Xu}]{Hatamizadeh2022UNETR}
\bibinfo{author}{Hatamizadeh, A.}, \bibinfo{author}{Tang, Y.}, \bibinfo{author}{Nath, V.}, \bibinfo{author}{Yang, D.}, \bibinfo{author}{Myronenko, A.}, \bibinfo{author}{Landman, B.}, \bibinfo{author}{Roth, H.R.}, \bibinfo{author}{Xu, D.}, \bibinfo{year}{2022}b.
\newblock \bibinfo{title}{{ UNETR: Transformers for 3D Medical Image Segmentation }} , \bibinfo{pages}{1748--1758}\DOIprefix\doi{10.1109/WACV51458.2022.00181}.
\bibitem[{He et~al.(2022)He, Chen, Xie, Li, Doll\'ar and Girshick}]{He2022MAE}
\bibinfo{author}{He, K.}, \bibinfo{author}{Chen, X.}, \bibinfo{author}{Xie, S.}, \bibinfo{author}{Li, Y.}, \bibinfo{author}{Doll\'ar, P.}, \bibinfo{author}{Girshick, R.}, \bibinfo{year}{2022}.
\newblock \bibinfo{title}{Masked autoencoders are scalable vision learners} , \bibinfo{pages}{16000--16009}.
\bibitem[{Horvat et~al.(2024)Horvat, Papanikolaou and Koh}]{Horvat2024RadiomicsLimitations}
\bibinfo{author}{Horvat, N.}, \bibinfo{author}{Papanikolaou, N.}, \bibinfo{author}{Koh, D.M.}, \bibinfo{year}{2024}.
\newblock \bibinfo{title}{Radiomics beyond the hype: A critical evaluation toward oncologic clinical use}.
\newblock \bibinfo{journal}{Radiology: Artificial Intelligence} \bibinfo{volume}{6}, \bibinfo{pages}{e230437}.
\newblock \DOIprefix\doi{10.1148/ryai.230437}.
\bibitem[{Hossain et~al.(2023)Hossain, Gul, Chowdhury, Khan, Sumon, Bhuiyan, Khandakar, Hossain, Sadique, Al-Hashimi, Ayari, Mahmud and Alqahtani}]{Hossain2023LiverSpleenAmbiguity}
\bibinfo{author}{Hossain, M.S.A.}, \bibinfo{author}{Gul, S.}, \bibinfo{author}{Chowdhury, M.E.H.}, \bibinfo{author}{Khan, M.S.}, \bibinfo{author}{Sumon, M.S.I.}, \bibinfo{author}{Bhuiyan, E.H.}, \bibinfo{author}{Khandakar, A.}, \bibinfo{author}{Hossain, M.}, \bibinfo{author}{Sadique, A.}, \bibinfo{author}{Al-Hashimi, I.}, \bibinfo{author}{Ayari, M.A.}, \bibinfo{author}{Mahmud, S.}, \bibinfo{author}{Alqahtani, A.}, \bibinfo{year}{2023}.
\newblock \bibinfo{title}{Deep learning framework for liver segmentation from t1-weighted mri images}.
\newblock \bibinfo{journal}{Sensors} \bibinfo{volume}{23}.
\newblock \DOIprefix\doi{10.3390/s23218890}.
\bibitem[{Isensee et~al.(2021)Isensee, Jaeger, Kohl, Petersen and Maier-Hein}]{Isensee2021nnUNet}
\bibinfo{author}{Isensee, F.}, \bibinfo{author}{Jaeger, P.F.}, \bibinfo{author}{Kohl, S.A.A.}, \bibinfo{author}{Petersen, J.}, \bibinfo{author}{Maier-Hein, K.H.}, \bibinfo{year}{2021}.
\newblock \bibinfo{title}{nnu-net: a self-configuring method for deep learning-based biomedical image segmentation}.
\newblock \bibinfo{journal}{Nature Methods} \bibinfo{volume}{18}, \bibinfo{pages}{203--211}.
\newblock \DOIprefix\doi{10.1038/s41592-020-01008-z}.
\bibitem[{Isensee et~al.(2025)Isensee, Rokuss, Krämer, Dinkelacker, Ravindran, Stritzke, Hamm, Wald, Langenberg, Ulrich, Deissler, Floca and Maier-Hein}]{Isensee2025nnInteractive}
\bibinfo{author}{Isensee, F.}, \bibinfo{author}{Rokuss, M.}, \bibinfo{author}{Krämer, L.}, \bibinfo{author}{Dinkelacker, S.}, \bibinfo{author}{Ravindran, A.}, \bibinfo{author}{Stritzke, F.}, \bibinfo{author}{Hamm, B.}, \bibinfo{author}{Wald, T.}, \bibinfo{author}{Langenberg, M.}, \bibinfo{author}{Ulrich, C.}, \bibinfo{author}{Deissler, J.}, \bibinfo{author}{Floca, R.}, \bibinfo{author}{Maier-Hein, K.}, \bibinfo{year}{2025}.
\newblock \bibinfo{title}{nninteractive: Redefining 3d promptable segmentation}.
\newblock \bibinfo{journal}{arXiv preprint arXiv:2503.08373} .
\bibitem[{Ishwaran et~al.(2008)Ishwaran, Kogalur, Blackstone and Lauer}]{Ishwaran2008RandomSurvivalForests}
\bibinfo{author}{Ishwaran, H.}, \bibinfo{author}{Kogalur, U.}, \bibinfo{author}{Blackstone, E.}, \bibinfo{author}{Lauer, M.}, \bibinfo{year}{2008}.
\newblock \bibinfo{title}{Random survival forests}.
\newblock \bibinfo{journal}{The Annals of Applied Statistics} \bibinfo{volume}{2}.
\newblock \DOIprefix\doi{10.1214/08-AOAS169}.
\bibitem[{Kirienko et~al.(2021)Kirienko, Sollini, Corbetta, Voulaz, Gozzi, Interlenghi, Gallivanone, Castiglioni, Asselta, Duga, Sold{\`a} and Chiti}]{Kirienko2021LogTransform}
\bibinfo{author}{Kirienko, M.}, \bibinfo{author}{Sollini, M.}, \bibinfo{author}{Corbetta, M.}, \bibinfo{author}{Voulaz, E.}, \bibinfo{author}{Gozzi, N.}, \bibinfo{author}{Interlenghi, M.}, \bibinfo{author}{Gallivanone, F.}, \bibinfo{author}{Castiglioni, I.}, \bibinfo{author}{Asselta, R.}, \bibinfo{author}{Duga, S.}, \bibinfo{author}{Sold{\`a}, G.}, \bibinfo{author}{Chiti, A.}, \bibinfo{year}{2021}.
\newblock \bibinfo{title}{Radiomics and gene expression profile to characterise the disease and predict outcome in patients with lung cancer}.
\newblock \bibinfo{journal}{European Journal of Nuclear Medicine and Molecular Imaging} \bibinfo{volume}{48}, \bibinfo{pages}{3643--3655}.
\newblock \DOIprefix\doi{10.1007/s00259-021-05371-7}.
\bibitem[{Kirillov et~al.(2023)Kirillov, Mintun, Ravi, Mao, Rolland, Gustafson, Xiao, Whitehead, Berg, Lo, Dollár and Girshick}]{Kirillov2023SAM}
\bibinfo{author}{Kirillov, A.}, \bibinfo{author}{Mintun, E.}, \bibinfo{author}{Ravi, N.}, \bibinfo{author}{Mao, H.}, \bibinfo{author}{Rolland, C.}, \bibinfo{author}{Gustafson, L.}, \bibinfo{author}{Xiao, T.}, \bibinfo{author}{Whitehead, S.}, \bibinfo{author}{Berg, A.C.}, \bibinfo{author}{Lo, W.Y.}, \bibinfo{author}{Dollár, P.}, \bibinfo{author}{Girshick, R.}, \bibinfo{year}{2023}.
\newblock \bibinfo{title}{Segment anything}.
\newblock \bibinfo{journal}{arXiv preprint arXiv:2304.02643} .
\bibitem[{Klein et~al.(2010)Klein, Staring, Murphy, Viergever and Pluim}]{Klein2010Elastix}
\bibinfo{author}{Klein, S.}, \bibinfo{author}{Staring, M.}, \bibinfo{author}{Murphy, K.}, \bibinfo{author}{Viergever, M.A.}, \bibinfo{author}{Pluim, J.P.W.}, \bibinfo{year}{2010}.
\newblock \bibinfo{title}{elastix: A toolbox for intensity-based medical image registration}.
\newblock \bibinfo{journal}{IEEE Transactions on Medical Imaging} \bibinfo{volume}{29}, \bibinfo{pages}{196--205}.
\newblock \DOIprefix\doi{10.1109/TMI.2009.2035616}.
\bibitem[{Kuo et~al.(2015)Kuo, Huang, Chiang, Yeh, Chan, Chen and Yu}]{Kuo2015TumorLocation}
\bibinfo{author}{Kuo, I.M.}, \bibinfo{author}{Huang, S.F.}, \bibinfo{author}{Chiang, J.M.}, \bibinfo{author}{Yeh, C.Y.}, \bibinfo{author}{Chan, K.M.}, \bibinfo{author}{Chen, J.S.}, \bibinfo{author}{Yu, M.C.}, \bibinfo{year}{2015}.
\newblock \bibinfo{title}{Clinical features and prognosis in hepatectomy for colorectal cancer with centrally located liver metastasis}.
\newblock \bibinfo{journal}{World Journal of Surgical Oncology} \bibinfo{volume}{13}, \bibinfo{pages}{92}.
\newblock \DOIprefix\doi{10.1186/s12957-015-0497-6}.
\bibitem[{Lambin et~al.(2012)Lambin, Rios-Velazquez, Leijenaar, Carvalho, {van Stiphout}, Granton, Zegers, Gillies, Boellard, Dekker and Aerts}]{Lambin2012Radiomics}
\bibinfo{author}{Lambin, P.}, \bibinfo{author}{Rios-Velazquez, E.}, \bibinfo{author}{Leijenaar, R.}, \bibinfo{author}{Carvalho, S.}, \bibinfo{author}{{van Stiphout}, R.G.}, \bibinfo{author}{Granton, P.}, \bibinfo{author}{Zegers, C.M.}, \bibinfo{author}{Gillies, R.}, \bibinfo{author}{Boellard, R.}, \bibinfo{author}{Dekker, A.}, \bibinfo{author}{Aerts, H.J.}, \bibinfo{year}{2012}.
\newblock \bibinfo{title}{Radiomics: Extracting more information from medical images using advanced feature analysis}.
\newblock \bibinfo{journal}{European Journal of Cancer} \bibinfo{volume}{48}, \bibinfo{pages}{441--446}.
\newblock \DOIprefix\doi{10.1016/j.ejca.2011.11.036}.
\bibitem[{Liu et~al.(2021)Liu, Lin, Cao, Hu, Wei, Zhang, Lin and Guo}]{Liu2021SwinTransformer}
\bibinfo{author}{Liu, Z.}, \bibinfo{author}{Lin, Y.}, \bibinfo{author}{Cao, Y.}, \bibinfo{author}{Hu, H.}, \bibinfo{author}{Wei, Y.}, \bibinfo{author}{Zhang, Z.}, \bibinfo{author}{Lin, S.}, \bibinfo{author}{Guo, B.}, \bibinfo{year}{2021}.
\newblock \bibinfo{title}{Swin transformer: Hierarchical vision transformer using shifted windows} , \bibinfo{pages}{9992--10002}\DOIprefix\doi{10.1109/ICCV48922.2021.00986}.
\bibitem[{Liyanage et~al.(2020)Liyanage, Zois and Chelmis}]{Liyanage2020JointSelectClassif}
\bibinfo{author}{Liyanage, Y.W.}, \bibinfo{author}{Zois, D.}, \bibinfo{author}{Chelmis, C.}, \bibinfo{year}{2020}.
\newblock \bibinfo{title}{On–the–fly feature selection and classification with application to civic engagement platforms} , \bibinfo{pages}{3762--3766}\DOIprefix\doi{10.1109/ICASSP40776.2020.9053564}.
\bibitem[{Ma et~al.(2024)Ma, He, Li, Han, You and Wang}]{Ma2024MedSAM}
\bibinfo{author}{Ma, J.}, \bibinfo{author}{He, Y.}, \bibinfo{author}{Li, F.}, \bibinfo{author}{Han, L.}, \bibinfo{author}{You, C.}, \bibinfo{author}{Wang, B.}, \bibinfo{year}{2024}.
\newblock \bibinfo{title}{Segment anything in medical images}.
\newblock \bibinfo{journal}{Nature Communications} \bibinfo{volume}{15}, \bibinfo{pages}{654}.
\newblock \DOIprefix\doi{10.1038/s41467-024-44824-z}.
\bibitem[{Ma et~al.(2025)Ma, Yang, Kim, Chen, Baharoon, Fallahpour, Asakereh, Lyu and Wang}]{Ma2025MedSAM2}
\bibinfo{author}{Ma, J.}, \bibinfo{author}{Yang, Z.}, \bibinfo{author}{Kim, S.}, \bibinfo{author}{Chen, B.}, \bibinfo{author}{Baharoon, M.}, \bibinfo{author}{Fallahpour, A.}, \bibinfo{author}{Asakereh, R.}, \bibinfo{author}{Lyu, H.}, \bibinfo{author}{Wang, B.}, \bibinfo{year}{2025}.
\newblock \bibinfo{title}{Medsam2: Segment anything in 3d medical images and videos}.
\newblock \bibinfo{journal}{arXiv preprint arXiv:2504.03600} .
\bibitem[{Mann and Whitney(1947)}]{Mann1947WilcoxonRankSum}
\bibinfo{author}{Mann, H.B.}, \bibinfo{author}{Whitney, D.R.}, \bibinfo{year}{1947}.
\newblock \bibinfo{title}{{On a Test of Whether one of Two Random Variables is Stochastically Larger than the Other}}.
\newblock \bibinfo{journal}{The Annals of Mathematical Statistics} \bibinfo{volume}{18}, \bibinfo{pages}{50 -- 60}.
\newblock \DOIprefix\doi{10.1214/aoms/1177730491}.
\bibitem[{Margonis and Vauthey(2022)}]{Margonis2022PrecisionSurgery}
\bibinfo{author}{Margonis, G.A.}, \bibinfo{author}{Vauthey, J.N.}, \bibinfo{year}{2022}.
\newblock \bibinfo{title}{Precision surgery for colorectal liver metastases: Current knowledge and future perspectives}.
\newblock \bibinfo{journal}{Annals of Gastroenterological Surgery} \bibinfo{volume}{6}, \bibinfo{pages}{606--615}.
\newblock \DOIprefix\doi{10.1002/ags3.12591}.
\bibitem[{Mariotti et~al.(2025)Mariotti, Agostini, Borgheresi, Marchegiani, Zannotti, Giacomelli, Pierpaoli, Tola, Galiffa and Giovagnoni}]{Mariotti2025RadiomicsReview}
\bibinfo{author}{Mariotti, F.}, \bibinfo{author}{Agostini, A.}, \bibinfo{author}{Borgheresi, A.}, \bibinfo{author}{Marchegiani, M.}, \bibinfo{author}{Zannotti, A.}, \bibinfo{author}{Giacomelli, G.}, \bibinfo{author}{Pierpaoli, L.}, \bibinfo{author}{Tola, E.}, \bibinfo{author}{Galiffa, E.}, \bibinfo{author}{Giovagnoni, A.}, \bibinfo{year}{2025}.
\newblock \bibinfo{title}{Insights into radiomics: a comprehensive review for beginners}.
\newblock \bibinfo{journal}{Clinical and Translational Oncology} \DOIprefix\doi{10.1007/s12094-025-03939-5}.
\bibitem[{Moor et~al.(2023)Moor, Banerjee, Abad, Krumholz, Leskovec, Topol and Rajpurkar}]{Moor2023FoundationModels}
\bibinfo{author}{Moor, M.}, \bibinfo{author}{Banerjee, O.}, \bibinfo{author}{Abad, Z.S.H.}, \bibinfo{author}{Krumholz, H.M.}, \bibinfo{author}{Leskovec, J.}, \bibinfo{author}{Topol, E.J.}, \bibinfo{author}{Rajpurkar, P.}, \bibinfo{year}{2023}.
\newblock \bibinfo{title}{Foundation models for generalist medical artificial intelligence}.
\newblock \bibinfo{journal}{Nature} \bibinfo{volume}{616}, \bibinfo{pages}{259--265}.
\newblock \DOIprefix\doi{10.1038/s41586-023-05881-4}.
\bibitem[{Nakai et~al.(2020)Nakai, Gonoi, Kurokawa, Nishioka, Abe, Arita, Ushiku, Hasegawa and Abe}]{Nakai2020CRLMPeripheral}
\bibinfo{author}{Nakai, Y.}, \bibinfo{author}{Gonoi, W.}, \bibinfo{author}{Kurokawa, R.}, \bibinfo{author}{Nishioka, Y.}, \bibinfo{author}{Abe, H.}, \bibinfo{author}{Arita, J.}, \bibinfo{author}{Ushiku, T.}, \bibinfo{author}{Hasegawa, K.}, \bibinfo{author}{Abe, O.}, \bibinfo{year}{2020}.
\newblock \bibinfo{title}{Mri findings of liver parenchyma peripheral to colorectal liver metastasis: A potential predictor of long-term prognosis}.
\newblock \bibinfo{journal}{Radiology} \bibinfo{volume}{297}, \bibinfo{pages}{584--594}.
\newblock \DOIprefix\doi{10.1148/radiol.2020202367}.
\bibitem[{Peng et~al.(2005)Peng, Long and Ding}]{Peng2005mRMR}
\bibinfo{author}{Peng, H.}, \bibinfo{author}{Long, F.}, \bibinfo{author}{Ding, C.}, \bibinfo{year}{2005}.
\newblock \bibinfo{title}{Feature selection based on mutual information criteria of max-dependency, max-relevance, and min-redundancy}.
\newblock \bibinfo{journal}{IEEE Transactions on Pattern Analysis and Machine Intelligence} \bibinfo{volume}{27}, \bibinfo{pages}{1226--1238}.
\newblock \DOIprefix\doi{10.1109/TPAMI.2005.159}.
\bibitem[{Ravi et~al.(2024)Ravi, Gabeur, Hu, Hu, Ryali, Ma, Khedr, Rädle, Rolland, Gustafson, Mintun, Pan, Alwala, Carion, Wu, Girshick, Dollár and Feichtenhofer}]{Ravi2024SAM2}
\bibinfo{author}{Ravi, N.}, \bibinfo{author}{Gabeur, V.}, \bibinfo{author}{Hu, Y.T.}, \bibinfo{author}{Hu, R.}, \bibinfo{author}{Ryali, C.}, \bibinfo{author}{Ma, T.}, \bibinfo{author}{Khedr, H.}, \bibinfo{author}{Rädle, R.}, \bibinfo{author}{Rolland, C.}, \bibinfo{author}{Gustafson, L.}, \bibinfo{author}{Mintun, E.}, \bibinfo{author}{Pan, J.}, \bibinfo{author}{Alwala, K.V.}, \bibinfo{author}{Carion, N.}, \bibinfo{author}{Wu, C.Y.}, \bibinfo{author}{Girshick, R.}, \bibinfo{author}{Dollár, P.}, \bibinfo{author}{Feichtenhofer, C.}, \bibinfo{year}{2024}.
\newblock \bibinfo{title}{Sam 2: Segment anything in images and videos}.
\newblock \bibinfo{journal}{arXiv preprint arXiv:2408.00714} .
\bibitem[{Rawla et~al.(2019)Rawla, Sunkara and Barsouk}]{Rawla2019CRC}
\bibinfo{author}{Rawla, P.}, \bibinfo{author}{Sunkara, T.}, \bibinfo{author}{Barsouk, A.}, \bibinfo{year}{2019}.
\newblock \bibinfo{title}{Epidemiology of colorectal cancer: incidence, mortality, survival, and risk factors}.
\newblock \bibinfo{journal}{Gastroenterology Review/Przegląd Gastroenterologiczny} \bibinfo{volume}{14}, \bibinfo{pages}{89--103}.
\newblock \DOIprefix\doi{10.5114/pg.2018.81072}.
\bibitem[{Rayed et~al.(2024)Rayed, Islam, Niha, Jim, Kabir and Mridha}]{Rayed2024DLLimitations}
\bibinfo{author}{Rayed, M.E.}, \bibinfo{author}{Islam, S.S.}, \bibinfo{author}{Niha, S.I.}, \bibinfo{author}{Jim, J.R.}, \bibinfo{author}{Kabir, M.M.}, \bibinfo{author}{Mridha, M.}, \bibinfo{year}{2024}.
\newblock \bibinfo{title}{Deep learning for medical image segmentation: State-of-the-art advancements and challenges}.
\newblock \bibinfo{journal}{Informatics in Medicine Unlocked} \bibinfo{volume}{47}, \bibinfo{pages}{101504}.
\newblock \DOIprefix\doi{10.1016/j.imu.2024.101504}.
\bibitem[{Roberts et~al.(2014)Roberts, White, Cockbain, Hodson, Hidalgo, Toogood and Lodge}]{Roberts2014BiomarkersLimitations}
\bibinfo{author}{Roberts, K.J.}, \bibinfo{author}{White, A.}, \bibinfo{author}{Cockbain, A.}, \bibinfo{author}{Hodson, J.}, \bibinfo{author}{Hidalgo, E.}, \bibinfo{author}{Toogood, G.J.}, \bibinfo{author}{Lodge, J.P.A.}, \bibinfo{year}{2014}.
\newblock \bibinfo{title}{Performance of prognostic scores in predicting long-term outcome following resection of colorectal liver metastases}.
\newblock \bibinfo{journal}{British Journal of Surgery} \bibinfo{volume}{101}, \bibinfo{pages}{856--866}.
\newblock \DOIprefix\doi{10.1002/bjs.9471}.
\bibitem[{Ronneberger et~al.(2015)Ronneberger, Fischer and Brox}]{Ronneberger2015UNet}
\bibinfo{author}{Ronneberger, O.}, \bibinfo{author}{Fischer, P.}, \bibinfo{author}{Brox, T.}, \bibinfo{year}{2015}.
\newblock \bibinfo{title}{U-net: Convolutional networks for biomedical image segmentation} , \bibinfo{pages}{234--241}.
\bibitem[{Sasaki et~al.(2018)Sasaki, Morioka, Conci, Margonis, Sawada, Ruzzenente, Kumamoto, Iacono, Andreatos, Guglielmi, Endo and Pawlik}]{Sasaki2018TBS}
\bibinfo{author}{Sasaki, K.}, \bibinfo{author}{Morioka, D.}, \bibinfo{author}{Conci, S.}, \bibinfo{author}{Margonis, G.A.}, \bibinfo{author}{Sawada, Y.}, \bibinfo{author}{Ruzzenente, A.}, \bibinfo{author}{Kumamoto, T.}, \bibinfo{author}{Iacono, C.}, \bibinfo{author}{Andreatos, N.}, \bibinfo{author}{Guglielmi, A.}, \bibinfo{author}{Endo, I.}, \bibinfo{author}{Pawlik, T.M.}, \bibinfo{year}{2018}.
\newblock \bibinfo{title}{The tumor burden score: A new ``metro-ticket'' prognostic tool for colorectal liver metastases based on tumor size and number of tumors}.
\newblock \bibinfo{journal}{Annals of Surgery} \bibinfo{volume}{267}.
\bibitem[{Scapicchio et~al.(2021)Scapicchio, Gabelloni, Barucci, Cioni, Saba and Neri}]{Scapicchio2021DeepRadiomics}
\bibinfo{author}{Scapicchio, C.}, \bibinfo{author}{Gabelloni, M.}, \bibinfo{author}{Barucci, A.}, \bibinfo{author}{Cioni, D.}, \bibinfo{author}{Saba, L.}, \bibinfo{author}{Neri, E.}, \bibinfo{year}{2021}.
\newblock \bibinfo{title}{A deep look into radiomics}.
\newblock \bibinfo{journal}{La radiologia medica} \bibinfo{volume}{126}, \bibinfo{pages}{1296--1311}.
\newblock \DOIprefix\doi{10.1007/s11547-021-01389-x}.
\bibitem[{Sengupta et~al.(2025)Sengupta, Chakrabarty and Soni}]{Sengupta2025SAMvsSAM2}
\bibinfo{author}{Sengupta, S.}, \bibinfo{author}{Chakrabarty, S.}, \bibinfo{author}{Soni, R.}, \bibinfo{year}{2025}.
\newblock \bibinfo{title}{Is sam 2 better than sam in medical image segmentation?} , \bibinfo{pages}{97}\DOIprefix\doi{10.1117/12.3047370}.
\bibitem[{SETH et~al.(2021)SETH, AMEMIYA, CHEUNG, HSIEH, LAW and MILOT}]{Seth2021MutationCRLM}
\bibinfo{author}{SETH, A.}, \bibinfo{author}{AMEMIYA, Y.}, \bibinfo{author}{CHEUNG, H.}, \bibinfo{author}{HSIEH, E.}, \bibinfo{author}{LAW, C.}, \bibinfo{author}{MILOT, L.}, \bibinfo{year}{2021}.
\newblock \bibinfo{title}{Delayed mri enhancement of colorectal cancer liver metastases is associated with metastatic mutational profile}.
\newblock \bibinfo{journal}{Cancer Genomics \& Proteomics} \bibinfo{volume}{18}, \bibinfo{pages}{627--635}.
\newblock \DOIprefix\doi{10.21873/cgp.20285}.
\bibitem[{Shur et~al.(2021)Shur, Doran, Kumar, ap~Dafydd, Downey, O’Connor, Papanikolaou, Messiou, Koh and Orton}]{Shur2021RadiomicsOncology}
\bibinfo{author}{Shur, J.}, \bibinfo{author}{Doran, S.}, \bibinfo{author}{Kumar, S.}, \bibinfo{author}{ap~Dafydd, D.}, \bibinfo{author}{Downey, K.}, \bibinfo{author}{O’Connor, J.P.B.}, \bibinfo{author}{Papanikolaou, N.}, \bibinfo{author}{Messiou, C.}, \bibinfo{author}{Koh, D.M.}, \bibinfo{author}{Orton, M.R.}, \bibinfo{year}{2021}.
\newblock \bibinfo{title}{Radiomics in oncology: A practical guide}.
\newblock \bibinfo{journal}{RadioGraphics} \bibinfo{volume}{41}, \bibinfo{pages}{1717--1732}.
\newblock \DOIprefix\doi{10.1148/rg.2021210037}. \bibinfo{note}{pMID: 34597235}.
\bibitem[{Spolverato et~al.(2013)Spolverato, Ejaz, Azad and Pawlik}]{Spolverato2013SurgicalOutcome}
\bibinfo{author}{Spolverato, G.}, \bibinfo{author}{Ejaz, A.}, \bibinfo{author}{Azad, N.}, \bibinfo{author}{Pawlik, T.M.}, \bibinfo{year}{2013}.
\newblock \bibinfo{title}{Surgery for colorectal liver metastases: The evolution of determining prognosis}.
\newblock \bibinfo{journal}{World J Gastrointest Oncol} \bibinfo{volume}{5}, \bibinfo{pages}{207--221}.
\newblock \DOIprefix\doi{10.4251/wjgo.v5.i12.207}.
\bibitem[{Sung et~al.(2021)Sung, Ferlay, Siegel, Laversanne, Soerjomataram, Jemal and Bray}]{Sung2021GLOBOCAN}
\bibinfo{author}{Sung, H.}, \bibinfo{author}{Ferlay, J.}, \bibinfo{author}{Siegel, R.L.}, \bibinfo{author}{Laversanne, M.}, \bibinfo{author}{Soerjomataram, I.}, \bibinfo{author}{Jemal, A.}, \bibinfo{author}{Bray, F.}, \bibinfo{year}{2021}.
\newblock \bibinfo{title}{Global cancer statistics 2020: Globocan estimates of incidence and mortality worldwide for 36 cancers in 185 countries}.
\newblock \bibinfo{journal}{CA: A Cancer Journal for Clinicians} \bibinfo{volume}{71}, \bibinfo{pages}{209--249}.
\newblock \DOIprefix\doi{10.3322/caac.21660}.
\bibitem[{Tang et~al.(2022)Tang, Yang, Li, Roth, Landman, Xu, Nath and Hatamizadeh}]{Tang2022FoundationSwin}
\bibinfo{author}{Tang, Y.}, \bibinfo{author}{Yang, D.}, \bibinfo{author}{Li, W.}, \bibinfo{author}{Roth, H.R.}, \bibinfo{author}{Landman, B.}, \bibinfo{author}{Xu, D.}, \bibinfo{author}{Nath, V.}, \bibinfo{author}{Hatamizadeh, A.}, \bibinfo{year}{2022}.
\newblock \bibinfo{title}{Self-supervised pre-training of swin transformers for 3d medical image analysis} , \bibinfo{pages}{20698--20708}\DOIprefix\doi{10.1109/CVPR52688.2022.02007}.
\bibitem[{Tian et~al.(2024)Tian, Wang, Wen, Wang, Li and Liu}]{Tian2024PrognosticFactors}
\bibinfo{author}{Tian, Y.}, \bibinfo{author}{Wang, Y.}, \bibinfo{author}{Wen, N.}, \bibinfo{author}{Wang, S.}, \bibinfo{author}{Li, B.}, \bibinfo{author}{Liu, G.}, \bibinfo{year}{2024}.
\newblock \bibinfo{title}{Prognostic factors associated with early recurrence following liver resection for colorectal liver metastases: a systematic review and meta-analysis}.
\newblock \bibinfo{journal}{BMC Cancer} \bibinfo{volume}{24}, \bibinfo{pages}{426}.
\newblock \DOIprefix\doi{10.1186/s12885-024-12162-4}.
\bibitem[{Ulrich et~al.(2023)Ulrich, Isensee, Wald, Zenk, Baumgartner and Maier-Hein}]{Ulrich2023MultiTalent}
\bibinfo{author}{Ulrich, C.}, \bibinfo{author}{Isensee, F.}, \bibinfo{author}{Wald, T.}, \bibinfo{author}{Zenk, M.}, \bibinfo{author}{Baumgartner, M.}, \bibinfo{author}{Maier-Hein, K.H.}, \bibinfo{year}{2023}.
\newblock \bibinfo{title}{Multitalent: A multi-dataset approach to medical image segmentation} , \bibinfo{pages}{648--658}.
\bibitem[{Vaswani et~al.(2017)Vaswani, Shazeer, Parmar, Uszkoreit, Jones, Gomez, Kaiser and Polosukhin}]{Vaswani2017attention}
\bibinfo{author}{Vaswani, A.}, \bibinfo{author}{Shazeer, N.}, \bibinfo{author}{Parmar, N.}, \bibinfo{author}{Uszkoreit, J.}, \bibinfo{author}{Jones, L.}, \bibinfo{author}{Gomez, A.N.}, \bibinfo{author}{Kaiser, L.u.}, \bibinfo{author}{Polosukhin, I.}, \bibinfo{year}{2017}.
\newblock \bibinfo{title}{Attention is all you need} \bibinfo{volume}{30}.
\bibitem[{Wang et~al.(2023)Wang, Zhong, Hu, Huang, Chen, Chen, Zhao, Wei and Li}]{Wang2023CRLM}
\bibinfo{author}{Wang, Y.}, \bibinfo{author}{Zhong, Xinyangand~He, X.}, \bibinfo{author}{Hu, Z.}, \bibinfo{author}{Huang, H.}, \bibinfo{author}{Chen, J.}, \bibinfo{author}{Chen, K.}, \bibinfo{author}{Zhao, S.}, \bibinfo{author}{Wei, P.}, \bibinfo{author}{Li, D.}, \bibinfo{year}{2023}.
\newblock \bibinfo{title}{Liver metastasis from colorectal cancer: pathogenetic development, immune landscape of the tumour microenvironment and therapeutic approaches}.
\newblock \bibinfo{journal}{Journal of Experimental {\&} Clinical Cancer Research} \bibinfo{volume}{42}, \bibinfo{pages}{177}.
\newblock \DOIprefix\doi{10.1186/s13046-023-02729-7}.
\bibitem[{Woerner and Baumgartner(2025)}]{Woerner2025ZeroVsFewShot}
\bibinfo{author}{Woerner, S.}, \bibinfo{author}{Baumgartner, C.F.}, \bibinfo{year}{2025}.
\newblock \bibinfo{title}{Navigating data scarcity using foundation models: A benchmark of few-shot and zero-shot learning approaches in medical imaging} , \bibinfo{pages}{30--39}.
\bibitem[{Xu et~al.(2025)Xu, Hosseini, Anderson, Rinaldi, Krishnan, Martel and Goubran}]{xu20253DINO}
\bibinfo{author}{Xu, T.}, \bibinfo{author}{Hosseini, S.}, \bibinfo{author}{Anderson, C.}, \bibinfo{author}{Rinaldi, A.}, \bibinfo{author}{Krishnan, R.G.}, \bibinfo{author}{Martel, A.L.}, \bibinfo{author}{Goubran, M.}, \bibinfo{year}{2025}.
\newblock \bibinfo{title}{A generalizable 3d framework and model for self-supervised learning in medical imaging}.
\newblock \bibinfo{journal}{arXiv preprint arXiv:2501.11755} .
\bibitem[{Zhao et~al.(2024)Zhao, Alzubaidi, Zhang, Duan and Gu}]{Zhao2024TransferLearning}
\bibinfo{author}{Zhao, Z.}, \bibinfo{author}{Alzubaidi, L.}, \bibinfo{author}{Zhang, J.}, \bibinfo{author}{Duan, Y.}, \bibinfo{author}{Gu, Y.}, \bibinfo{year}{2024}.
\newblock \bibinfo{title}{A comparison review of transfer learning and self-supervised learning: Definitions, applications, advantages and limitations}.
\newblock \bibinfo{journal}{Expert Systems with Applications} \bibinfo{volume}{242}, \bibinfo{pages}{122807}.
\newblock \DOIprefix\doi{10.1016/j.eswa.2023.122807}.
\bibitem[{Zhou et~al.(2022a)Zhou, He, Ma, Berg-Kirkpatrick and Neubig}]{Zhou2022PromptGeneralization}
\bibinfo{author}{Zhou, C.}, \bibinfo{author}{He, J.}, \bibinfo{author}{Ma, X.}, \bibinfo{author}{Berg-Kirkpatrick, T.}, \bibinfo{author}{Neubig, G.}, \bibinfo{year}{2022}a.
\newblock \bibinfo{title}{Prompt consistency for zero-shot task generalization} , \bibinfo{pages}{2613--2626}\DOIprefix\doi{10.18653/v1/2022.findings-emnlp.192}.
\bibitem[{Zhou et~al.(2022b)Zhou, Liu, Wang, Wen, Amador, Yuan, Ran, Xiong, Ran, Chen and Wen}]{Zhou2022LiverMetastasis}
\bibinfo{author}{Zhou, H.}, \bibinfo{author}{Liu, Z.}, \bibinfo{author}{Wang, Y.}, \bibinfo{author}{Wen, X.}, \bibinfo{author}{Amador, E.H.}, \bibinfo{author}{Yuan, L.}, \bibinfo{author}{Ran, X.}, \bibinfo{author}{Xiong, L.}, \bibinfo{author}{Ran, Y.}, \bibinfo{author}{Chen, W.}, \bibinfo{author}{Wen, Y.}, \bibinfo{year}{2022}b.
\newblock \bibinfo{title}{Colorectal liver metastasis: molecular mechanism and interventional therapy}.
\newblock \bibinfo{journal}{Signal Transduction and Targeted Therapy} \bibinfo{volume}{7}, \bibinfo{pages}{70}.
\newblock \DOIprefix\doi{10.1038/s41392-022-00922-2}.
\bibitem[{Zhu et~al.(2024)Zhu, Hamdi, Qi, Jin and Wu}]{Zhu2024MedicalSAM2}
\bibinfo{author}{Zhu, J.}, \bibinfo{author}{Hamdi, A.}, \bibinfo{author}{Qi, Y.}, \bibinfo{author}{Jin, Y.}, \bibinfo{author}{Wu, J.}, \bibinfo{year}{2024}.
\newblock \bibinfo{title}{Medical sam 2: Segment medical images as video via segment anything model 2}.
\newblock \bibinfo{journal}{arXiv preprint arXiv:2408.00874} .

\end{thebibliography}
\end{document}